\documentclass[letterpaper]{article} 
\usepackage[preprint]{aaai2027}  
\usepackage[hyphens]{url}  
\usepackage{graphicx} 
\usepackage{natbib}  
\usepackage{caption} 
\usepackage{algorithm}
\usepackage{algorithmic}

\usepackage{newfloat}
\usepackage{listings}
\DeclareCaptionStyle{ruled}{labelfont=normalfont,labelsep=colon,strut=off} 
\floatstyle{ruled}
\newfloat{listing}{tb}{lst}{}
\floatname{listing}{Listing}

\usepackage{booktabs}
\usepackage{array}

\usepackage{amsmath}
\usepackage{amssymb}
\usepackage{mathtools}

\newtheorem{lemma}{Lemma}

\usepackage{xcolor}

\title{How Temporal Correlations Shape Memory in Linear Recurrent Neural Networks}
\author{
    Arnol Manuel Fokam\textsuperscript{\rm 1}\corresponding,
    Fasseu Sieyondji Akpevwoghene\textsuperscript{\rm 2},
    Edem Fiifi Dawson\textsuperscript{\rm 3}
}
\affiliations{
    \textsuperscript{\rm 1}Independent Researcher, United Kingdom\\
    \textsuperscript{\rm 2}Department of Electrical and Electronic Engineering, University of Buea, Cameroon\\
    \textsuperscript{\rm 3}minoHealth AI Labs, Ghana\\
    arnolfokam23@gmail.com, dfasseu18@gmail.com, edem.fiifi.dawson@gmail.com
}

\ifdefined\pdfinfoomitdate\pdfinfoomitdate=1\fi
\ifdefined\pdfsuppressptexinfo\pdfsuppressptexinfo=-1\fi
\ifdefined\pdftrailerid\pdftrailerid{}\fi

\begin{document}

\maketitle

\begin{abstract}
The linear recurrent neural network (LRNN) is a simple model for studying how much memory a network builds up as it trains. For uncorrelated inputs, earlier work found that training itself settles the network between keeping the past and reacting only to the present. Real sequences are correlated, and we solve the learning dynamics exactly for correlated inputs. In the solution, keeping the past carries a cost. The whole effect of correlation lands on that cost. This cost reduces to the earlier one when inputs are uncorrelated and grows once they are positively correlated. Three findings follow. (1) Correlation reshapes the course of learning, not only its end. Memory builds, overshoots, and is partly removed, and the settled network keeps less of the past. (2) Memory switches off at a threshold set by one number, how much each input resembles the one just before it. Neither sequence length nor longer-range correlation moves this threshold. Memory is worth keeping only when the task needs the previous input more than the current input already supplies it through correlation with the past. (3) The best network changes too. Zero error demands a feedthrough, a path that passes the current input straight to the network's output and remembers nothing, and training builds it unprompted when given one spare hidden dimension. Our work turns one property of the input into a prediction of whether a network learns memory and explains why correlated data turns recurrent networks into change detectors.
\end{abstract}

\section{Introduction}
Recurrent neural networks (RNNs) appear across machine learning and neuroscience, as models that solve tasks with temporal structure and as models of neural dynamics themselves. Recent theory has begun to describe how training shapes them \citep{saxe2014exact,proca2025learning}, but no tractable account yet says how a network decides what to remember
when its inputs are correlated.

A key limitation of existing theoretical frameworks is their reliance on temporally independent inputs \citep{smekal2024towards,proca2025learning,bordelon2025dynamics}. Yet real-world data are rarely independent, instead exhibiting temporal correlations across domains such as language, physics, and neuroscience. By assuming temporally independent inputs, existing theory cannot account for the interaction between input predictability and memory formation. As input predictability increases, the benefit of retaining past information diminishes, suggesting an inherent trade-off between external statistical structure and internally stored memory. A quantitative theory describing this trade-off is still lacking.

To overcome this challenge, we consider linear RNNs as an analytically tractable model of recurrent computation. Although simplified, linear RNNs retain the essential mechanisms governing temporal information propagation while admitting exact mathematical analysis. Their close relationship to modern state-space models (SSMs), such as S4 \citep{gu2022efficiently} and Mamba \citep{gu2023mamba}, further makes them an appropriate framework for developing theoretical insights applicable to contemporary sequence modeling.

The linear RNN framework of \citet{proca2025learning} showed that for temporally independent inputs training itself sets a memory dial, settling the network between keeping the past and reacting only to the present. In this work we extend that framework to temporally correlated inputs and derive an exact theory of how input predictability shapes the memory a network learns.

\paragraph{Our contributions.}
\begin{enumerate}
\item We solve the learning dynamics of a linear RNN under correlated
inputs. The whole effect of correlation lands on one term, the price
the network pays for holding onto the past
(Lemma~\ref{lem:energy}).

\item We prove when that price is worth paying. One comparison decides
it, the task's demand for the previous input against the correlation
of the input's two most recent steps, and both sides can be read off a
dataset before training starts (Lemma~\ref{lem:boundary}).

\item We derive the network the task actually demands. A single
recurrent mode cannot be it. What it lacks is a feedthrough that skips
the recurrence entirely, and training recruits one as soon as the
network has a dimension to spare.
\end{enumerate} 

\section{Mathematical Setup}

\subsection{The Task}

The task is sequence-to-one regression, learned from a set of $P$ training
trajectories
\[
\big\{\big(x_{p,1}, \dots, x_{p,T},\, y_p\big)\big\}_{p=1}^{P}.
\]
Trajectory $p$ pairs a sequence of $T$ inputs $x_{p,i} \in \mathbb{R}^{N_x}$,
arriving one step at a time, with a target $y_p \in \mathbb{R}^{N_y}$. After
the last input the network produces a single output matching that target.

The task is specified by two families of statistics. The first is the
sequence of matrices
\[
\Sigma^{YX_i} = \frac{1}{P}\sum_{p=1}^{P} y_p\, x_{p,i}^{\top},
\qquad i = 1, \dots, T,
\]
the average correlation between the inputs at step $i$ and their targets,
one matrix per step. Each says how strongly the target leans on that step's
input. The second is the inputs' temporal structure, carried by the
input--input correlation
\[
\Sigma^{X_iX_j} = \frac{1}{P}\sum_{p=1}^{P} x_{p,i}\, x_{p,j}^{\top},
\qquad i, j = 1, \dots, T,
\]
one matrix per pair of timesteps, saying how much the input at step $i$
resembles the input at step $j$.

\textbf{Assumption~1 (constant singular vectors):} Every input--output
correlation matrix has the same singular vectors and differs only in its
singular values, such that $\Sigma^{YX_i} = U S_i V^{\top}$ with
$S_i = \mathrm{diag}(s_{\alpha,i})$, where $\alpha = 1, \dots, N$ runs over
the $N$ singular directions. These singular directions are the
independent directions of the task, one demand profile $s_{\alpha,i}$
per direction.

\textbf{Assumption~2 (whitened features):} Within one timestep the input
features are uncorrelated and whitened across all dimensions, such that
$\Sigma^{X_iX_i} = I$ for all $i$.

\textbf{Assumption~3 (temporally correlated inputs):} Across timesteps the
inputs may be correlated, such that $\Sigma^{X_iX_j} = C_{ij}\, I$; the
covariance is stationary, $C_{ij} = C_{\lvert i-j\rvert}$, the same for every
pair of steps $i$ and $j$ with the same gap $\lvert i-j\rvert$.

Assumption~3 is where this paper departs from earlier work on learning
dynamics in linear recurrent networks
\citep{smekal2024towards,proca2025learning,bordelon2025dynamics}. Their
whitened-input assumption
makes two demands at once. It wants white features within a timestep,
$\Sigma^{X_iX_i} = I$, and no correlation across timesteps,
$\Sigma^{X_iX_j} = 0$ for $i \neq j$. We keep only the
first demand as our Assumption~2. Instead of forbidding correlation across
time our Assumption~3 lets the inputs carry a temporal covariance $C$.
Correlation is the realistic case for sequence tasks. The relations across
time that make a task need memory are exactly what previous work set aside
by assuming uncorrelated inputs.

\subsection{The Architecture}

The LRNN reads a trajectory one input at a time, keeps a running hidden
state $h_i \in \mathbb{R}^{N_h}$, and produces a single output $\hat{y}$ when
the sequence ends. It has three weight matrices.
$W_x \in \mathbb{R}^{N_h \times N_x}$ reads each input into the state,
$W_h \in \mathbb{R}^{N_h \times N_h}$ carries the state forward one step, and
$W_y \in \mathbb{R}^{N_y \times N_h}$ reads the final state out,
\begin{equation}
h_{i+1} = W_h h_i + W_x x_i, \qquad \hat{y} = W_y h_{T+1},
\label{eq:model}
\end{equation}
starting from $h_1 = 0$.

Unrolling the recurrence writes the output as a weighted sum over the inputs,
\begin{equation}
\hat{y} = \sum_{i=1}^{T} M_i\, x_i, \qquad M_i = W_y W_h^{T-i} W_x.
\label{eq:unrolled}
\end{equation}
An input at step $i$ passes through the recurrence $T - i$ times before it
reaches the output, so the further back an input sits, the more copies of
$W_h$ it goes through.

\subsection{The Objective Function}

Training minimizes the squared error between the output and the target,
averaged over the $P$ trajectories,
\begin{equation}
\mathcal{L} = \frac{1}{2P} \sum_{p=1}^{P} \Big\lVert\, y_p - W_y \sum_{i=1}^{T} W_h^{T-i} W_x\, x_{p,i} \,\Big\rVert^2.
\label{eq:loss}
\end{equation}
We train the LRNN by gradient flow, the small-step limit of gradient
descent. Each weight matrix follows
\begin{equation}
\tau\, \frac{dW}{dt_\theta} = -\frac{\partial \mathcal{L}}{\partial W},
\qquad W \in \{W_x, W_h, W_y\},
\label{eq:flow}
\end{equation}
with $\tau$ the learning timescale and $t_\theta$ the training time, a
clock separate from the sequence position $i$.

\textbf{Assumption~4 (aligned initialization):} At initialization, training
time $t_\theta = 0$, the model is aligned to the task's singular vectors $U$
and $V$ of Assumption~1, taking the form
\[
W_x = V_h\,\mathrm{diag}(a_\alpha)\,V^{\top}, \qquad
W_h = V_h\,\mathrm{diag}(b_\alpha)\,V_h^{\top},
\]
\[
W_y = U\,\mathrm{diag}(c_\alpha)\,V_h^{\top}.
\]
The hidden basis $V_h$ has one column per task direction and
orthonormal columns, $V_h^{\top} V_h = I$.

Assumptions~1 and~4 are the standard conditions that make the learning
dynamics solvable in closed form \citep{saxe2014exact,proca2025learning}.

\section{Findings}

Our findings follow from one calculation. We differentiate the loss
\eqref{eq:loss}, evaluate its trajectory averages under the correlated
inputs of Assumption~3 in place of the white inputs of earlier work,
and follow the gradient flow \eqref{eq:flow} that results. The data
enters the dynamics only through the singular values of Assumption~1
and the covariance of Assumption~3, and everything below is read off
these two statistics. The derivations, the machine
verification, and the real-data validation are given in
Appendices~A through~E, provided as supplementary material.

\begin{figure*}[t]
\centering
\includegraphics[width=\textwidth]{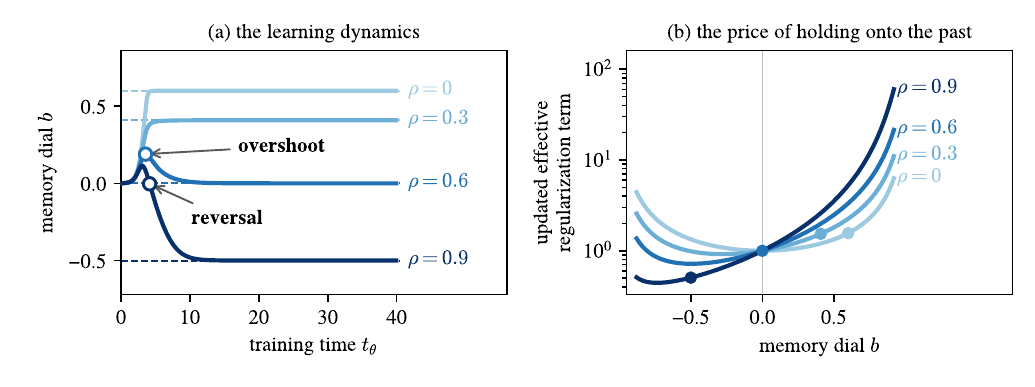}
\caption{Correlation raises the price of holding onto the past, and
the learning dynamics settle where the new price is worth paying.
Both panels use the geometric task profile $s_i = \lambda^{T-i}$ with
$\lambda = 0.6$ and an AR(1) input covariance
$C_{ij} = \rho^{\lvert i-j\rvert}$, at $T = 20$; darker curves carry
stronger correlation $\rho$. \textbf{(a)} The memory dial under
training. Solid curves are the simulated gradient flow of
Eqs.~\eqref{eq:flow-a}--\eqref{eq:flow-c} from small weights; dashed
lines are the energy minima, and every run reaches its minimum. The
white-input run settles at the task demand $b = \lambda$, and $\rho =
0.3$ drags the settle point down to $0.41$. At $\rho = \lambda$ the
dial overshoots (circle) before returning to zero, and at $\rho =
0.9$ it reverses (circle) and settles at $-0.5$, storing change
rather than level. \textbf{(b)} The updated effective regularization
term of Lemma~\ref{lem:energy},
$\sum_{i,j} b^{T-i} C_{ij}\, b^{T-j}$, against the memory dial.
Raising $\rho$ makes positive memory strictly dearer and change
detection ($b < 0$) cheaper; dots mark the $b$ each run settles on.}
\label{fig:dynamics-price}
\end{figure*}

\paragraph{Simulation details.} We integrate
\eqref{eq:flow-a}--\eqref{eq:flow-c} by explicit Euler at $\tau = 1$
with step $10^{-3}$ in training time, from the aligned small-weight
start $a = c = 0.05$ and $b = 0$, to $t_\theta = 40$ for
Figure~\ref{fig:dynamics-price} and $t_\theta = 30$ for
Figure~\ref{fig:grid-dynamics}. Figure~\ref{fig:recruit-grid} instead
trains the full weight matrices of \eqref{eq:model} from random
weights of scale $0.05$, twenty networks per cell, by the same scheme
with step $4 \times 10^{-3}$ to $t_\theta = 400$. The flow reaches the
data only through the two families of statistics, so every run here
uses those exactly, draws no finite sample of trajectories, and
carries no sampling error. Every run is
integrated to a fixed horizon rather than stopped adaptively, and each
settled dial reported here matches the corresponding minimum of the
energy \eqref{eq:energy} to three decimals.

\subsection{Correlation Reshapes the Learning Dynamics}

The learning dynamics of the Mathematical Setup have three
components. The first
is what changes, the three weight matrices of the architecture
\eqref{eq:model}, $W_x$, $W_h$, and $W_y$. The second is what pushes
them, the loss \eqref{eq:loss}, built from the task statistics of
Assumptions~1--3. The third is the rule that turns the push into
motion, the gradient flow \eqref{eq:flow}. In general these dynamics are
hard to follow. The matrices multiply one another in every term of the
unrolled map \eqref{eq:unrolled}, and the recurrence enters through its
first $T$ powers. The loss is therefore a high-dimensional non-convex
landscape over all the weight entries at once. Under Assumptions~1--4 the
problem breaks into independent pieces, one per task direction (the
singular directions of Assumption~1). The gradient flow keeps the pieces
separate (derivation in Appendix~A).

The problem along one direction is to learn just three scalars, its
\emph{connectivity modes} $a$, $b$, $c$ \citep{proca2025learning}. Here $a$ is the scale
of the input weights $W_x$, $b$ the scale of the recurrence $W_h$, and
$c$ the scale of the output weights $W_y$. The input at step $i$
reaches the output through the single path $c\, b^{T-i} a$, passing
through the recurrence $T - i$ times, so $b$ is the memory dial.

We fix one task direction $\alpha$ and suppress its index, writing
$s_i$ for $s_{\alpha,i}$. Along this direction the gradient flow
\eqref{eq:flow} becomes one differential equation per scalar,
\begin{align}
\tau\, \frac{da}{dt_\theta} &= \sum_{i=1}^{T} c b^{T-i}
\big( s_i - c a \textstyle\sum\limits_{j=1}^{T} C_{ij} b^{T-j} \big),
\label{eq:flow-a} \\
\tau\, \frac{db}{dt_\theta} &= \sum_{i=1}^{T-1} (T-i) c b^{T-i-1} a
\big( s_i - c a \textstyle\sum\limits_{j=1}^{T} C_{ij} b^{T-j} \big),
\label{eq:flow-b} \\
\tau\, \frac{dc}{dt_\theta} &= \sum_{i=1}^{T} b^{T-i} a
\big( s_i - c a \textstyle\sum\limits_{j=1}^{T} C_{ij} b^{T-j} \big).
\label{eq:flow-c}
\end{align}

The parenthesis shared by the three
equations is the residual error at step $i$. Its two terms are the
same measurement taken on two different signals, the correlation with
the input at step $i$ averaged over the training trajectories. The
first term $s_i$ is the \emph{required correlation}, the correlation
the target has with that input, the singular value of Assumption~1
read along this direction. The second term
$c\, a \sum_{j=1}^{T} C_{ij}\, b^{T-j}$
is the \emph{produced correlation}, the correlation the network's own
output has with that input. It sums the response to every input $j$,
the path $c\, b^{T-j} a$, scaled by the overlap $C_{ij}$ between
inputs $j$ and $i$. Every weight update
is proportional to their difference, and learning stops once the
produced correlation matches the required one at every step.

The overlap $C_{ij}$ is the whole effect of correlation on the
updates, and it enters only through the produced correlation.
Intuitively,
each input partly repeats the inputs it correlates with, so one
response covers several steps at once. Less memory is then needed to
meet the required correlations, and training stops with smaller
weights. Correlation acts as free
memory. It can even overshoot. When the produced correlation at a
step exceeds the required one, the residual error reverses sign and
the updates remove memory instead of building it, pushing the memory
dial $b$ toward zero and below. So of the three components of the
dynamics, correlation touches only the push. The quantities being
learned and the rule they follow are untouched; it is the score the
data assigns the network's output that changes, and through that
score correlation moves both the path training takes and the weights
it settles on. Figure~\ref{fig:dynamics-price}a shows these
dynamics: as correlation rises the dial settles lower, overshoots
and returns to zero, and finally reverses below it.

Where the dial settles can be read off one function. The three
equations collapse into gradient flow on an \emph{energy} $E$, the
direction's share of the loss \eqref{eq:loss}. The following lemma
gives $E$ in closed form; Appendix~B proves it from the
per-direction reduction of Appendix~A and checks that its $C = I$
specialization collapses to the white-input energy of
\citet{proca2025learning}.

\begin{lemma}[the per-mode energy]
\label{lem:energy}
Given Assumptions~1--4, the energy governing each task direction
decomposes into a data-driven term and an effective regularization
term,
\begin{multline}
E = \underbrace{\frac{1}{2} \sum_{i=1}^{T}
\Big( s_i^2 - 2\, s_i\, c\, b^{T-i} a \Big)}_{\text{data-driven term}} \\
+ \underbrace{\frac{1}{2}\, c^2 a^2
\sum_{i,j=1}^{T} b^{T-i}\, C_{ij}\, b^{T-j}}_{\text{updated effective regularization term}},
\label{eq:energy}
\end{multline}
the direction's share of the loss \eqref{eq:loss} up to an
additive constant independent of the weights.
\end{lemma}

The two terms are the two forces on the connectivity modes. The
data-driven term pulls the response at each step toward the required
correlation $s_i$. The effective regularization term is the price of
output itself, the mean square of the direction's output on the
actual inputs, and it is the one term the overlap $C_{ij}$ reshapes.

Earlier work on white inputs already found that remembering is
costly \citep{proca2025learning}. With $C = I$ the regularization sum
of \eqref{eq:energy} keeps only its diagonal, and each diagonal term
grows as the memory dial $b$ rises, so holding the past by itself
inflates the penalty.
Lemma~\ref{lem:energy} places the input covariance inside that sum,
and the covariance sets the new price through the cross terms
$b^{T-i}\, C_{ij}\, b^{T-j}$ with $i \neq j$. When the inputs are
positively correlated and $b > 0$, every cross term adds to the
penalty, so memory costs strictly more than in the white case. When
$b < 0$ the powers alternate in sign, the same cross terms can
cancel, and a mode that stores change rather than level pays less.
Figure~\ref{fig:dynamics-price}b traces this penalty against the
dial for increasing correlation.

\subsection{Correlation Decides When Memory Matters}

The curves of Figure~\ref{fig:dynamics-price}a end in three ways.
Most settle at a positive memory, lower as the correlation rises.
The run at $\rho = \lambda = 0.6$ builds a little memory and then
removes all of it; there the input already supplies all the past the
task demands. And the strongest correlation drives the dial below
zero into change detection. This finding shows the three endings are
predictable before training. One comparison, the task's one-step
demand against the correlation of the input's two most recent steps,
decides which ending training delivers.

Whether memory grows at all is decided at the memoryless setting
$b = 0$. There the network
reads only the last input, and the first trace of memory holds only
the previous input, so only the two most recent steps can vote.
In the energy \eqref{eq:energy} the modes $a$ and $c$ appear only
through their product, the gain $g = c\, a$, which we hold at its
best value for each setting of the dial. Expanding the two sums of
the energy to first order in $b$ at that best gain gives two
competing terms. The data-driven term grows with the demand
$s_{T-1}$ and the regularization term with the overlap
$C_{T-1,T}$, each measured against its own value at zero memory,
the demand $s_T$ and the variance $C_{T,T}$. The sign of their
balance decides whether the dial leaves zero (derivation in
Appendix~C, which proves Lemma~\ref{lem:boundary} together with
its two invariances). 

\begin{lemma}[the memory boundary]
\label{lem:boundary}
Given Assumptions~1--4 and a task that reads the newest input,
$s_T > 0$, gradient flow from small weights leaves the memoryless
setting toward positive memory exactly when
\begin{equation}
\frac{s_{T-1}}{s_T} \;>\; \frac{C_{T-1,\,T}}{C_{T,\,T}},
\label{eq:boundary}
\end{equation}
and the dial settles at $b = 0$ on the boundary. The condition
depends on the task only through its one-step demand and on the
input only through the correlation of its two most recent steps; the
sequence length $T$ and every other entry of $C$ leave it unchanged.
\end{lemma}

The condition is proved at the memoryless setting itself, so it
locates where memory switches on. That this local boundary is also
the global one, with no distant setting of the dial overtaking it,
is confirmed by simulation in Appendix~C, which also gives the
curvature condition under which the release is continuous.

In words, memory pays only when the task needs the previous input
more than the current input already supplies it through correlation
with the past. A correlated input is free
memory. The present carries a copy of the recent past, the copy
covers part of the required correlations, and the network stores
only what the copy cannot supply. Below the boundary the same
balance pushes the dial negative, and the network detects change
instead of storing the past. For the geometric task and AR(1) input
of Figure~\ref{fig:dynamics-price} the boundary reads
$\lambda = \rho$.

\begin{figure*}[t]
\centering
\includegraphics[width=\textwidth]{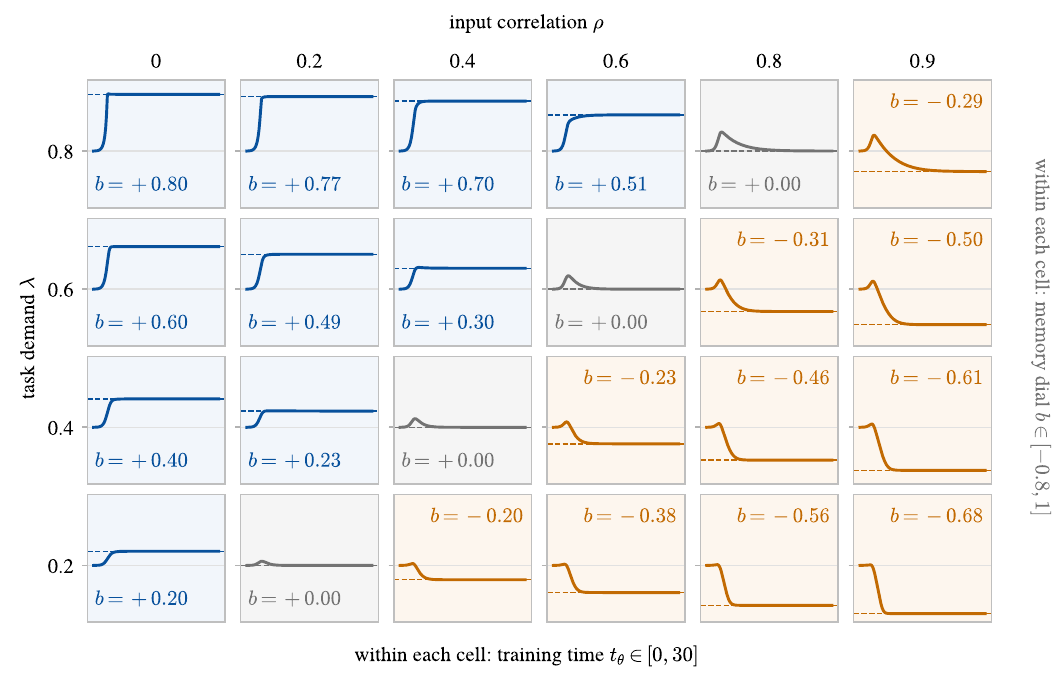}
\caption{One comparison decides every cell. Each cell trains a
network at one pairing of task demand $\lambda$ (rows) and input
correlation $\rho$ (columns) and shows the trajectory of the memory
dial $b$. Inputs are AR(1) with the geometric task at $T = 20$; the
shared per-cell axes are annotated on the grid's edges. Solid curves
are simulations, dashed lines the energy minima, and every run lands
on its minimum; the printed value is the settled $b$ and the shading
follows the outcome. Above the boundary ($\lambda > \rho$, blue) memory is built and kept, larger as
the demand outruns the correlation. On the boundary
(gray) the dial builds a little and
settles back to zero. Below it ($\lambda < \rho$, orange) training removes memory and the
mode turns to change detection, deeper as the imbalance grows.}
\label{fig:grid-dynamics}
\end{figure*}

Figure~\ref{fig:grid-dynamics} trains one network per cell over a
$4 \times 6$ grid of task demand and input correlation. Every cell
with $\lambda > \rho$ builds and keeps memory, every cell with
$\lambda = \rho$ settles back to zero, and every cell with
$\lambda < \rho$ reverses into change detection. The boundary of
Lemma~\ref{lem:boundary} is visible as the gray diagonal of the
grid. Appendix~C extends the sweep to a full $10 \times 10$ grid
(Figure~S1), where the same diagonal separates the
outcomes.

Earlier work could see only the leftmost column of
Figure~\ref{fig:grid-dynamics}. Its whitened inputs fix $\rho = 0$.
There the right-hand side of \eqref{eq:boundary} vanishes, any
demand on the past switches memory on, and only the task decides
\citep{proca2025learning}. The horizontal axis of the figure did not
exist in that theory, and every cell to the right of the first
column is opened by Assumption~3. Correlation lifts the boundary off
zero, and the task must now out-lean the data. The two invariances are what
make the law usable. Neither the sequence length nor any
correlation beyond the most recent pair moves the boundary, so a
single measurable statistic of a dataset fixes the threshold every
task must clear.

The law also holds on real data. Across nineteen series spanning
finance, weather, hydrology, brain recordings, and sunspots, the
learned crossing lands on the measured threshold. The full table, the task
construction, the protocol, and the scope condition that says which series the
law reaches are given in Appendix~E.

\begin{figure*}[t]
\centering
\includegraphics[width=\textwidth]{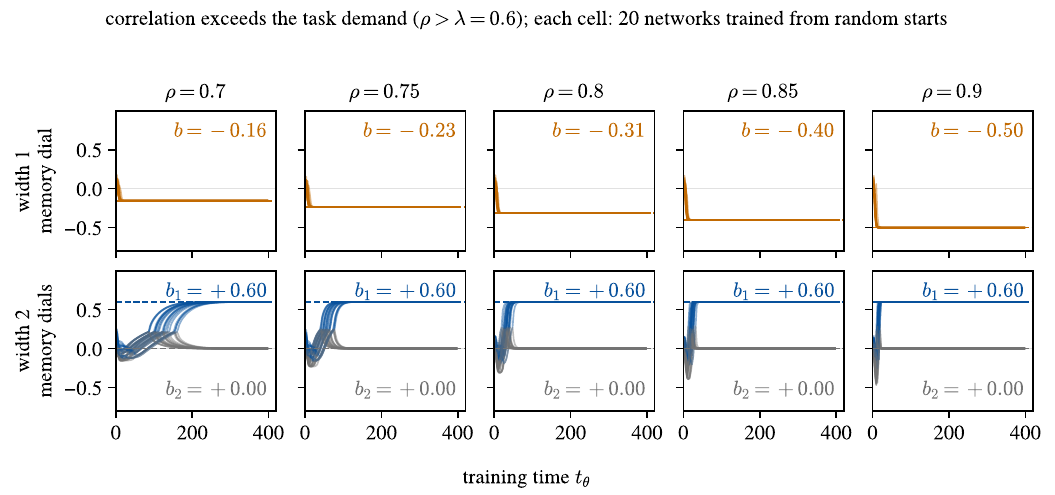}
\caption{Given one spare hidden dimension, training builds the
feedthrough on its own. Each cell pairs one width, the number of
hidden dimensions (rows), with one input correlation $\rho$
(columns); the five correlations all outrun the demand,
$\rho = 0.7$ to $0.9$ against the geometric task at
$\lambda = 0.6$, with AR(1) inputs at $T = 20$. Each cell trains
twenty networks of the unmodified model \eqref{eq:model} by
gradient flow from small random weights. Curves are the
trajectories of each hidden dimension's memory dial. At width one
the lone dial (orange) is dragged into change detection, settling
at $b = -0.16$, $-0.23$, $-0.31$, $-0.40$, $-0.50$ as $\rho$
grows. At width two the dial of the first hidden dimension (blue)
settles at the task demand $0.6$ and the second dial (gray) at
zero, both to three decimals in all one hundred runs.
The split comes faster the stronger the correlation.}
\label{fig:recruit-grid}
\end{figure*}

\subsection{Correlation Demands a Feedthrough Term}

Every run of Figure~\ref{fig:grid-dynamics} lands on the minimum of
its energy \eqref{eq:energy}, yet landing on the minimum is not the
same as solving the task. The minimum is the lowest point the three
connectivity modes can reach, while solving means reaching the
lowest error the task allows, and the two need not agree. In the
leftmost column they do. The dial stops at $b = \lambda$, every
required correlation is matched, and every residual error is zero.
In the columns to the right they come apart. The one exception is
the gray diagonal, where the dial at zero solves the task exactly.
Elsewhere the settled network leaves required correlations
unmatched at every step, and no setting of $a$, $b$, $c$ closes
the distance.

Closing the distance begins with writing down what solving demands.
Solving means the residual error of
\eqref{eq:flow-a}--\eqref{eq:flow-c} is zero at every step, a
produced correlation equal to the required one at each of the $T$
steps.

The produced correlation at step $i$ sums the network's weight on
every input $j$ through the overlaps $C_{ij}$, so zero residual
error at every step is a linear system, one equation per step, $T$
equations in all. A single hidden dimension brings three numbers
$a$, $b$, $c$ to $T$ demands. Write $w_j$ for the weight the
network places on the input at step $j$; in a single hidden
dimension this is the path $c\, b^{T-j} a$. Free the weights to
ask what the task alone allows, and the system and its solution
read
\begin{equation}
\sum_{j=1}^{T} C_{ij}\, w_j = s_i,
\quad i = 1, \dots, T,
\qquad
w^{*} = C^{-1} s,
\label{eq:wiener}
\end{equation}
with $s$ the vector collecting the required correlations.
Classical linear estimation knows $w^{*}$ as the Wiener filter
\citep{kailath2000linear} (derivation in Appendix~D). For the
geometric task and AR(1) input of Figure~\ref{fig:grid-dynamics}
the solution has a readable shape, the same geometric \emph{tail}
a single hidden dimension already carries, at the task's own rate
$\lambda$, plus one extra \emph{spike} on the last input.

The present still carries the copy of the recent past, the copy
covers part of the required correlations, and the tail is scaled
down to store only what the copy cannot supply. A scaled-down tail
underweights the present itself, and the spike restores the
present's own weight, larger as the correlation grows. The tail
gain vanishes exactly on the gray diagonal, where the spike alone
is the whole solution and the dial at zero delivers it, the
exception of the opening paragraph. At any setting of its dial a
single hidden dimension is one \emph{ray}, a geometric profile
whose weight on the present is welded to its tail by the shared
gain $c\, a$; off the diagonal no ray at any rate equals a tail
plus a spike (proof in Appendix~D). Each hidden dimension of the
model \eqref{eq:model} carries a memory dial of its own, the
diagonal of $W_h$ in the basis training settles into (the
off-diagonal decays), and the network's
weight on each input is the sum of one ray per dimension. So zero
error demands a second hidden dimension with its dial at zero.

Whether training builds that second dimension on its own is the
next question. Figure~\ref{fig:recruit-grid} trains the unmodified
model \eqref{eq:model} from small random weights in place of the
aligned initialization of Assumption~4, at widths one and two,
where width counts hidden dimensions. Width two grants one
\emph{spare} dimension beyond the one the task's single direction
uses. At width one training
reverses the lone dial into change detection, deeper as the
correlation grows, and since no ray equals a tail plus a spike
every one of these cells settles above the lowest error the task
allows. At width two all hundred runs split identically, the first
dial to the task demand $0.6$ and the second to zero, both to
three decimals. The
deeper the correlation the sooner the split arrives, the spike
learned at a speed set by how badly it is needed.

With its dial at zero the spare dimension carries the spike; it
passes the last input straight to the output and remembers
nothing, precisely the direct-feedthrough $D$ term that
state-space models such as S4 \citep{gu2022efficiently}, S4D
\citep{Gu2022}, the LRU \citep{orvieto2023resurrecting}, and Mamba
\citep{gu2023mamba} attach beside their recurrence. When the copy
is built from more than the previous input the single spike widens
into free weights on the few most recent inputs, the short causal
convolution Mamba and H3 also carry \citep{fu2023hungry}. The
shape of what a single hidden dimension cannot supply is set by
the structure of the input's correlation, and an AR(1) input is
the case whose answer is a single feedthrough (Table~S1 in Appendix~D). Those components were engineering choices; under
correlated input they are what zero error demands, and given one
spare hidden dimension gradient flow builds the feedthrough
unprompted.

Appendix~D derives the filter \eqref{eq:wiener}, gives the tail
and spike gains in closed form with their sign behaviour, proves
that no single ray matches a tail plus a spike, and tabulates, one
covariance family per row, the component beyond a single ray that
zero error demands (Table~S1).

\section{Related Work}

\paragraph{Learning dynamics of linear networks.}
Our analysis builds on the exact-solution program for deep \emph{linear} networks \citep{saxe2014exact}, in which learning decouples along the singular directions of the input-output correlation into independent scalar modes. Later work extends this program to generalization, the rich and lazy regimes, and implicit bias toward low-rank solutions \citep{atanasovneural, gissin2020the, NEURIPS2023_17a9ab41}. \citet{proca2025learning} carry this program to linear recurrent networks and make the memory dial a learned quantity. Their per-mode energy is the object we generalize. The whole effect of temporal correlation lands on their memory cost, which becomes the regularization sum $\sum_{i,j=1}^{T} b^{T-i} C_{ij}\, b^{T-j}$ of Lemma~\ref{lem:energy} and recovers their theory at $C = I$. Two nearby works stop short of correlated inputs, \citet{smekal2024towards} solving deep state-space models with the recurrent spectrum held fixed and \citet{bordelon2025dynamics} tracking a single integrating eigenvalue and naming the correlated-input question as open. The architectures our feedthrough result speaks to, S4 \citep{gu2022efficiently}, the LRU \citep{orvieto2023resurrecting}, and Mamba \citep{gu2023mamba}, already carry the direct-feedthrough term it recovers.

\paragraph{Memory capacity of recurrent networks.}
A parallel line measures how much past a recurrent network can hold. Memory capacity was defined for fixed random reservoirs \citep{jaeger:techreport2002} and bounded for linear recurrent networks \citep{White2004}. The Fisher memory curve quantifies retention against noise \citep{Ganguli2008}, and the information-processing-capacity framework exposes a memory-nonlinearity trade-off \citep{Dambre2012}. That tradition fixes the network and asks what an architecture can store. We ask instead what memory training builds, and show it is set by the input's correlation rather than by capacity alone.

\paragraph{Optimal estimation under correlated inputs.}
Pricing memory under correlated inputs places the work in the classical theory of linear estimation. The optimal readout we derive is the Wiener filter for stationary signals \citep{Wiener1949}, equivalently the generalized-least-squares estimator \citep{kailath2000linear}, and in its state-space innovations form, whitening the input is the first step of optimal prediction \citep{Kalman1960}. What is new is not the filter but its placement: it is reached here by gradient descent under a recurrent capacity constraint. Which recurrence learning selects, and when memory collapses, are questions estimation theory does not pose.

\paragraph{Whitening in the brain.}
The rule our optimal recurrence follows, subtract the predictable part and keep only what is new, is the efficient-coding principle of sensory neuroscience, where the retina whitens temporally and spatially correlated input \citep{Srinivasan1982,6795763}. Our contribution is to show a recurrent network trained by gradient descent arrives at the same whitening, and to turn one measurable statistic of the data, the correlation of its two most recent steps $C_{T-1,\,T}/C_{T,\,T}$ from Lemma~\ref{lem:boundary}, into a prediction of how much memory the network will keep.

\section{Conclusion}

We asked how the temporal structure of data shapes the memory a
recurrent network learns. The answer is a price. Correlation enters the
learning dynamics in one place, the price a mode pays for holding onto
the past, and the rest follows from it. Memory is built when the task's
demand for the previous input outruns the correlation the input already
carries, and it is removed when it does not. That comparison needs one
number from the data, so the outcome can be read off a dataset before
training starts. Below the boundary the network does not simply store
less, it changes what it computes and turns into a change detector. And
the network that solves the task exactly is not a recurrence alone but
a recurrence beside a feedthrough, a component modern state-space
models already carry and gradient descent recruits on its own. Our
results do not make a recurrent network more accurate. They make it
predictable, which is what matching a sequence model to a problem
requires.


\renewcommand{\thefigure}{S\arabic{figure}}
\renewcommand{\thetable}{S\arabic{table}}
\renewcommand{\theequation}{S\arabic{equation}}
\setcounter{figure}{0}
\setcounter{table}{0}
\setcounter{equation}{0}

\appendix

\section{A. Mode Decoupling under Correlated Inputs}
\label{app:mode-decoupling}

There are two kinds of correlation, and only one would threaten decoupling. Correlation across time relates the input now to the input earlier. Correlation across modes relates direction $\alpha$ to a different direction $\beta$. Decoupling requires only that the second vanish. Under Assumption~3, the temporal covariance is isotropic in feature space, a scalar $C_{ij}$ times the identity, so it can carry structure across time but none across modes. Steps~1 and~2 verify this for the input and input-target statistics; Step~3 assembles the per-mode split of the loss; and Step~4 shows that gradient flow preserves the alignment assumed in Eq.~\eqref{eq:appA-aligned}.

\paragraph{Aligned representation.}
Assumption~4 initializes the weights in the task singular-vector bases:
\begin{align}
W_x &= V_h\operatorname{diag}(a_\alpha)V^\top,\nonumber\\
W_h &= V_h\operatorname{diag}(b_\alpha)V_h^\top,\nonumber\\
W_y &= U\operatorname{diag}(c_\alpha)V_h^\top.
\tag{A.1}
\label{eq:appA-aligned}
\end{align}
Here $V_h^\top V_h=I$. Contracting the adjacent hidden bases inside
$M_i=W_yW_h^{T-i}W_x$ gives
\begin{equation}
M_i
=U\operatorname{diag}\!\left(c_\alpha b_\alpha^{T-i}a_\alpha\right)V^\top.
\tag{A.2}
\label{eq:appA-per-step}
\end{equation}
Equation~\eqref{eq:appA-per-step} maps a single input step to the output. The network reads out
the whole sequence as $\widehat y=\sum_{i=1}^{T}M_ix_i$, one term per step. Rotate
$\widetilde y=U^\top y$ and $\widetilde x_i=V^\top x_i$, and substitute
Eq.~\eqref{eq:appA-per-step} into that sum. Every factor is diagonal, so the $\alpha$ coordinate of
the output collects only the $\alpha$ entry of each:
\begin{equation}
(U^\top\widehat y)_\alpha
=c_\alpha a_\alpha\sum_{i=1}^{T}b_\alpha^{T-i}
\widetilde x_{\alpha,i}.
\tag{A.3}
\label{eq:appA-output-coordinate}
\end{equation}
The exponent $T-i$ counts the recurrent steps between input $i$ and the readout,
so the newest input $i=T$ arrives undamped and the oldest is scaled by
$b_\alpha^{T-1}$.

\paragraph{Step 1: The input covariance remains diagonal across modes.}
The paper defines $\Sigma^{YX_i}$ and $\Sigma^{X_iX_j}$ as averages over the $P$
training trajectories. We write $\mathbb E$ for the population second moments those
averages estimate, impose Assumptions~1--3 on the population objects, and reserve
$\langle\cdot\rangle_P$ for the finite sample. By Assumption~3,
\[
\Sigma^{X_iX_j}
= \mathbb E[x_ix_j^\top]
= C_{ij}I.
\tag{A.4}
\label{eq:appA-input-cov}
\]
Since $\widetilde x_i=V^\top x_i$ with $V$ fixed by Assumption~1, linearity of expectation gives
\begin{align*}
\mathbb E[\widetilde x_i\widetilde x_j^\top]
&=V^\top\mathbb E[x_ix_j^\top]V
 =V^\top\Sigma^{X_iX_j}V,\\
V^\top\Sigma^{X_iX_j}V
&=V^\top(C_{ij}I)V
 =C_{ij}(V^\top V)
 =C_{ij}I.
\end{align*}
Reading off the $(\alpha,\beta)$ entry gives
\[
\mathbb E[\widetilde x_{\alpha,i}\widetilde x_{\beta,j}]
= C_{ij}\delta_{\alpha\beta}.
\tag{A.5}
\label{eq:supp-appA-input-statistic}
\]
The four-index expectation has factored into a temporal part and a mode part. The temporal part $C_{ij}$ is unconstrained; the mode part $\delta_{\alpha\beta}$ is as sparse as a factor can be. What forces that sparsity is the identity in Eq.~\eqref{eq:appA-input-cov}: a scalar multiple of $I$ survives any orthogonal change of basis, so no rotation can manufacture a correlation between modes, however strong the temporal structure.

\paragraph{Step 2: The input-target statistic is diagonal in the same modes.}
Both rotations are fixed by Assumption~1, so they pass through the expectation:
\begin{align*}
\mathbb E[\widetilde y\widetilde x_i^\top]
&=\mathbb E[U^\top yx_i^\top V]\\
&=U^\top\mathbb E[yx_i^\top]V
 =U^\top\Sigma^{YX_i}V.
\end{align*}
Assumption~1 also supplies the decomposition $\Sigma^{YX_i}=US_iV^\top$ in these
same bases. Substituting it lets each rotation cancel against its own transpose:
\begin{align*}
U^\top\Sigma^{YX_i}V
&=U^\top(US_iV^\top)V\\
&=(U^\top U)S_i(V^\top V)
 =S_i.
\end{align*}
Here $S_i=\operatorname{diag}(s_{\alpha,i})$ is diagonal, again by Assumption~1. Reading off the $(\alpha,\beta)$ entry yields
\[
\mathbb E[\widetilde y_\alpha\widetilde x_{\beta,i}]
= s_{\alpha,i}\delta_{\alpha\beta}.
\tag{A.6}
\label{eq:appA-task-statistic}
\]
Like Eq.~\eqref{eq:supp-appA-input-statistic}, this statistic carries structure across timesteps and none across modes. The quantity $s_{\alpha,i}$ is the task profile of mode $\alpha$, the target's demand on the input at step $i$.

\paragraph{Step 3: The loss splits into one problem per mode.}
Let $\langle f\rangle_P=P^{-1}\sum_{p=1}^P f_p$ denote the trajectory average. Equation~\eqref{eq:loss} is
\[
\begin{aligned}
L
&=\frac12\left\langle\lVert y-\widehat y\rVert^2\right\rangle_P\\
&=\mathrm{const}+\frac12\sum_\alpha
\left\langle
\left(\widetilde y_\alpha-(U^\top\widehat y)_\alpha\right)^2
\right\rangle_P.
\end{aligned}
\tag{A.7}
\label{eq:appA-loss-split-start}
\]
The squared length is a sum of coordinatewise squares over the single mode index $\alpha$. The rotated loss equals the original up to a term independent of the weights, since the prediction lies in the column space of $U$ by Eq.~\eqref{eq:appA-per-step}; that term is absorbed by the constant carried through to Eq.~\eqref{eq:appA-mode-energy}. Equation~\eqref{eq:appA-output-coordinate} makes each $(U^\top\widehat y)_\alpha$ depend on mode $\alpha$ alone, so no cross-mode coupling can enter a summand.

Take one summand and substitute Eq.~\eqref{eq:appA-output-coordinate} for the prediction. Expanding the square gives three terms:
\[
\begin{aligned}
&\left(\widetilde y_\alpha-c_\alpha a_\alpha
 \sum_i b_\alpha^{T-i}\widetilde x_{\alpha,i}\right)^2\\
&\quad=\widetilde y_\alpha^2
 -2c_\alpha a_\alpha\,\widetilde y_\alpha
 \sum_i b_\alpha^{T-i}\widetilde x_{\alpha,i}\\
&\qquad+c_\alpha^2a_\alpha^2
 \left(\sum_i b_\alpha^{T-i}\widetilde x_{\alpha,i}\right)
 \left(\sum_j b_\alpha^{T-j}\widetilde x_{\alpha,j}\right).
\end{aligned}
\]
The last term is the same sum multiplied by itself. Renaming the second copy's index to $j$ turns the product into a double sum, in which the two weights combine as $b_\alpha^{T-i}b_\alpha^{T-j}=b_\alpha^{2T-i-j}$. Averaging over trajectories and moving $\langle\cdot\rangle_P$ inside each sum by linearity gives
\[
\begin{aligned}
&\left\langle
\left(\widetilde y_\alpha-(U^\top\widehat y)_\alpha\right)^2
\right\rangle_P\\
&\quad=\left\langle\widetilde y_\alpha^2\right\rangle_P\\
&\qquad-2c_\alpha a_\alpha\sum_i b_\alpha^{T-i}
 \left\langle\widetilde y_\alpha\widetilde x_{\alpha,i}\right\rangle_P\\
&\qquad+c_\alpha^2a_\alpha^2\sum_{i,j}b_\alpha^{2T-i-j}
 \left\langle\widetilde x_{\alpha,i}\widetilde x_{\alpha,j}\right\rangle_P.
\end{aligned}
\tag{A.8}
\]
Substituting Eqs.~\eqref{eq:supp-appA-input-statistic} and~\eqref{eq:appA-task-statistic}, the parameter-dependent terms of the mode-$\alpha$ loss are
\[
\begin{aligned}
E_\alpha(a_\alpha,b_\alpha,c_\alpha)
=\frac12\sum_i&\Bigl(s_{\alpha,i}^2
 -2s_{\alpha,i}c_\alpha b_\alpha^{T-i}a_\alpha\Bigr)\\
&+\frac12c_\alpha^2a_\alpha^2
 \sum_{i,j}b_\alpha^{T-i}C_{ij}b_\alpha^{T-j}.
\end{aligned}
\tag{A.9}
\label{eq:appA-mode-energy}
\]
up to a mode-dependent constant that is independent of the weights. Thus
\[
L=\mathrm{const}+\sum_\alpha E_\alpha(a_\alpha,b_\alpha,c_\alpha).
\tag{A.10}
\]
The covariance enters only through the second sum, the memory cost. Any difference between $\langle\widetilde y_\alpha^2\rangle_P$ and the convenient constant $\sum_is_{\alpha,i}^2$ leaves every gradient unchanged.

\paragraph{Step 4: Gradient flow preserves the alignment.}
Decoupling at initialization is not enough. We show that the gradient at every aligned point is again mode-diagonal and hence tangent to the aligned set. Write
\[
\begin{aligned}
\widehat y&=\sum_iM_ix_i=\sum_iA_iW_xx_i,\\
A_i&=W_yW_h^{T-i},\\
M_i&=A_iW_x.
\end{aligned}
\tag{A.11}
\]
Differentiating Eq.~\eqref{eq:appA-loss-split-start} with respect to $W_x$ gives
\[
\frac{\partial L}{\partial W_x}
=\sum_iA_i^\top
\left(\sum_jC_{ij}M_j-\Sigma^{YX_i}\right).
\tag{A.12}
\label{eq:appA-Wx-gradient}
\]
Gradient flow moves along the negative of this expression; its sign does not affect the shape argument. The inner sum follows because $\mathbb E[\widehat yx_i^\top]=\sum_jM_j\Sigma^{X_jX_i}=\sum_jC_{ji}M_j=\sum_jC_{ij}M_j$, while $\mathbb E[yx_i^\top]=\Sigma^{YX_i}$.

Throughout this step, $\operatorname{diag}(\cdot)$ denotes a diagonal over $\alpha$ whose entries need not be written explicitly. From Eq.~\eqref{eq:appA-aligned},
\begin{align*}
A_i
&=U\operatorname{diag}\!\left(c_\alpha b_\alpha^{T-i}\right)V_h^\top,\\
A_i^\top
&=V_h\operatorname{diag}\!\left(c_\alpha b_\alpha^{T-i}\right)U^\top.
\end{align*}
Both $M_j=U\operatorname{diag}(\cdot)V^\top$ and $\Sigma^{YX_i}=US_iV^\top$ have the same task-frame shape, so
\[
\sum_jC_{ij}M_j-\Sigma^{YX_i}
=U\operatorname{diag}(\cdot)V^\top.
\tag{A.13}
\]
Multiplying by $A_i^\top$ and contracting $U^\top U=I$ gives
\[
A_i^\top\!
\left(U\operatorname{diag}(\cdot)V^\top\right)
=V_h\operatorname{diag}(\cdot)V^\top.
\tag{A.14}
\]
Summing over $i$ preserves this shape. Hence the $W_x$ gradient has the same mode-diagonal form as $W_x$ in Eq.~\eqref{eq:appA-aligned}, and no cross-mode component is created.

The other gradients close in the same way. The $W_y$ gradient is
\[
\frac{\partial L}{\partial W_y}
=\sum_i
\left(\sum_jC_{ij}M_j-\Sigma^{YX_i}\right)
\left(W_h^{T-i}W_x\right)^\top,
\tag{A.15}
\]
and its factors contract to $U\operatorname{diag}(\cdot)V_h^\top$, the form of $W_y$. For $W_h$, differentiating the power produces the product-rule sum
\[
\sum_{k=0}^{T-i-1}W_h^k(\cdot)W_h^{T-i-1-k}.
\tag{A.16}
\]
Each term has the shape $V_h\operatorname{diag}(\cdot)V_h^\top$, the form of $W_h$. The extra sum over $k$ changes only the diagonal entries and cannot rotate them away from the $V_h$ basis.

All three gradients therefore preserve the aligned mode-diagonal forms in Eq.~\eqref{eq:appA-aligned}. A trajectory that starts aligned never develops a cross-mode component, and the full-matrix flow reduces to the per-mode flow of Appendix~B, one diagonal entry at a time. We also confirm this reduction numerically at random aligned points.

Steps~1 and~2 place all temporal structure on the time axis and none on the mode axis. Step~3 splits the loss, and Step~4 shows that gradient flow preserves the split. This proves the reduction used by the first Finding.

\section{B. Memory Cost and Per-Mode Learning Dynamics}
\label{app:memory-cost}

Two hypotheses hold throughout this appendix and Appendices~C--D, in addition to Assumptions~1--4 of the main paper. The covariance matrix $C=(C_{ij})$ is positive definite, which is what makes the profiled gain and the Wiener solution well defined. Each recurrent mode is taken in the stable range $\lvert b\rvert<1$, which is the domain on which the settled dial is read. Neither is needed for the main paper's statements; both are used only in the derivations here.

This appendix derives the equations used in the first Finding. Appendix~A allows us to fix one task mode and suppress its index. The only change from temporally white inputs is the variance of the mode's recurrent output, the regularization sum of Lemma~\ref{lem:energy}, which is a quadratic form in the memory weights,
\[
\begin{aligned}
\sum_{i,j=1}^{T}b^{T-i}C_{ij}b^{T-j}&=v(b)^\top Cv(b),\\
v(b)&=(b^{T-1},\ldots,b,1)^\top,
\end{aligned}
\tag{B.1}
\label{eq:appB-cost-vector}
\]
with $v$ the vector of memory weights $b^{T-i}$. With temporally white inputs the same sum keeps only its diagonal, the white cost $R_0$ of Eq.~\eqref{eq:appB-white-cost} below. The parameterization by $(a,b,c)$ is unchanged.

Two sums of Eq.~\eqref{eq:appA-mode-energy} recur in every derivation from here on. Neither is a new object: they are the two halves of the paper's Lemma~\ref{lem:energy}. The first is the memory-weighted demand $\sum_{i=1}^{T}s_ib^{T-i}$, the part of the data-driven term the dial moves; the second is the regularization sum $\sum_{i,j=1}^{T}b^{T-i}C_{ij}b^{T-j}$, what that memory costs. Both are differentiated below, so we record the two derivatives once:
\[
\begin{aligned}
\frac{d}{db}\sum_{i=1}^{T}s_ib^{T-i}
&=\sum_{i=1}^{T-1}(T-i)s_ib^{T-i-1},\\
\frac{d}{db}\sum_{i,j=1}^{T}b^{T-i}C_{ij}b^{T-j}
&=\sum_{i,j=1}^{T}(2T-i-j)b^{2T-i-j-1}C_{ij}.
\end{aligned}
\tag{B.2}
\label{eq:appB-reward-cost}
\]
The $i=T$ term of the first sum and the $i=j=T$ term of the second are constant in $b$ and drop. After profiling out the gain $g=ca$, the objective left for the dial is the data-driven term measured against the regularization term, and the dial climbs whenever the demand slope beats the overlap slope; at zero memory that comparison is $s_{T-1}/s_T$ against $C_{T,T-1}/C_{TT}$, the boundary of Lemma~\ref{lem:boundary}.

\paragraph{Per-mode starting point.}
Let $x_i=\widetilde x_{\alpha,i}$ and $y=\widetilde y_\alpha$ denote the rotated input and target for one fixed mode. The scalar prediction is
\[
\widehat y=ca\sum_{i=1}^{T}b^{T-i}x_i.
\tag{B.3}
\]
Its expected squared error, up to a constant independent of $(a,b,c)$, is
\begin{align}
E(a,b,c)
&=\frac12\sum_{i=1}^{T}\bigl(s_i^2-2s_icb^{T-i}a\bigr)
\nonumber\\
&\quad+\frac12c^2a^2\sum_{i,j=1}^{T}b^{T-i}C_{ij}b^{T-j},
\tag{B.4}
\label{eq:appB-energy}
\end{align}
the data-driven term first and the updated effective regularization term second. The factor $1/\tau$ belongs only to the gradient-flow timescale. Writing $\dot b$ for $db/dt_\theta$, differentiation gives the single flush-centered system
\begin{align}
\tau\dot a
&=\sum_{i=1}^{T}cb^{T-i}
\bigl(s_i-ca\textstyle\sum\limits_{j=1}^{T}C_{ij}b^{T-j}\bigr),
\nonumber\\
\tau\dot b
&=\sum_{i=1}^{T-1}(T-i)cb^{T-i-1}a
\bigl(s_i-ca\textstyle\sum\limits_{j=1}^{T}C_{ij}b^{T-j}\bigr),
\nonumber\\
\tau\dot c
&=\sum_{i=1}^{T}b^{T-i}a
\bigl(s_i-ca\textstyle\sum\limits_{j=1}^{T}C_{ij}b^{T-j}\bigr).
\tag{B.5}
\label{eq:appB-flow}
\end{align}
These equations conserve the balanced invariant $a^2-c^2$, the recurrent analogue of \citet{saxe2014exact}.

Eq.~\eqref{eq:appB-flow} is Eqs.~\eqref{eq:flow-a}--\eqref{eq:flow-c} of the main text. The $b$-equation collects the two halves of the double sum with the symmetry of $C$, which is why the factor $2T-i-j$ of Eq.~\eqref{eq:appB-reward-cost} becomes the single factor $T-i$. Their shared parenthesis is the step-$i$ residual
\[
s_i-ca\sum_jC_{ij}b^{T-j}.
\tag{B.6}
\]
The main paper reads this residual in words, and the two documents describe the
same objects under different names. The \emph{required correlation} is $s_i$,
the correlation the target has with the input at step $i$. The \emph{produced correlation}
is what is subtracted from it, $ca\sum_jC_{ij}b^{T-j}$, the correlation the
network's own output has with that same input; the residual is the amount by
which the second falls short of the first. The \emph{memory dial} is $b$.
Lemma~\ref{lem:energy}, the per-mode energy, is Eq.~\eqref{eq:appB-energy} above, and the
paper's \emph{updated effective regularization term} is its
$\tfrac12c^2a^2\sum_{i,j}b^{T-i}C_{ij}b^{T-j}$, the memory cost of
Eq.~\eqref{eq:appB-cost-vector}, the remaining terms being the
\emph{data-driven term}. Lemma~\ref{lem:boundary}, the memory boundary, is
Eq.~\eqref{eq:appC-margin} of Appendix~C, with $C_0=C_{TT}$ and
$C_1=C_{T,T-1}$; Eq.~\eqref{eq:appC-final-boundary} restates it in the $\lambda^\star$ notation of Appendix~C.
Where this document says only ``the dial'' or ``the memory cost,'' those are the
objects meant.
For temporally white inputs, $\mathbb E[x_ix_j]=\delta_{ij}$ and the
regularization sum collapses to
\[
R_0(b)=\sum_{i=1}^{T}b^{2T-2i}=\frac{1-b^{2T}}{1-b^2}.
\tag{B.7}
\label{eq:appB-white-cost}
\]
The memory-weighted demand and the parameterization are unchanged; only the memory cost must be recomputed.

The derivation has five steps. Step~1 shows that correlation changes the memory cost but not the data-driven term. Step~2 rewrites that cost as the quadratic form in Eq.~\eqref{eq:appB-cost-vector}. Step~3 isolates the correlation surcharge beyond the white cost. Step~4 profiles the gain out of the energy, leaving the demand measured against the memory cost. Step~5 matches the recurrent flow's interior fixed points to the critical points of that profiled energy.

\paragraph{Step 1: Correlation changes the memory cost, not the data-driven term.}
Assumption~2 fixes the unit diagonal and Assumption~3 supplies the off-diagonal entries, so together they replace the temporally white statistic by
\[
\mathbb E[x_ix_j]=C_{ij},
\qquad
C_{ii}=1.
\tag{B.8}
\]
The data-driven term depends on the trajectories only through $\mathbb E[yx_i]=s_i$ and is therefore unchanged:
\[
\mathbb E[y\widehat y]
=ca\sum_ib^{T-i}\mathbb E[yx_i]
=ca\sum_{i=1}^{T}s_ib^{T-i}.
\tag{B.9}
\]
All new content lies in $\mathbb E[\widehat y^2]$, the one term in which $\mathbb E[x_ix_j]$ enters.

\paragraph{Step 2: The memory cost is the quadratic form $v^\top C v$.}
Write the unit-gain memory output as
\[
z=\sum_{i=1}^{T}b^{T-i}x_i,
\qquad
\widehat y=ca z.
\tag{B.10}
\]
Its mean square is
\[
\begin{aligned}
\mathbb E[z^2]
&=\sum_{i,j=1}^{T}b^{2T-i-j}\mathbb E[x_ix_j]\\
&=\sum_{i,j=1}^{T}b^{2T-i-j}C_{ij}.
\end{aligned}
\tag{B.11}
\]
With $v(b)$ from Eq.~\eqref{eq:appB-cost-vector},
\[
\mathbb E[z^2]=v(b)^\top Cv(b)\geq0,
\tag{B.12}
\]
because $C\succeq0$. Putting this quadratic form where the white cost stood is the only change to the parameter-dependent part of the energy.

\paragraph{Step 3: The correlation surcharge makes the same positive memory cost more.}
Split the regularization sum of Eq.~\eqref{eq:appB-energy} into diagonal and off-diagonal contributions, using $C_{ii}=1$:
\[
\begin{aligned}
\sum_{i,j=1}^{T}b^{2T-i-j}C_{ij}
&=\underbrace{\sum_ib^{2T-2i}}_{R_0(b),\ \text{Proca et al.'s white cost}}\\
&\quad+\underbrace{\sum_{i\neq j}b^{2T-i-j}C_{ij}}_{\text{correlation surcharge}}.
\end{aligned}
\tag{B.13}
\label{eq:appB-surcharge}
\]
The first term is exactly the white cost in Eq.~\eqref{eq:appB-white-cost}. Take positively correlated inputs to mean $C_{ij}>0$ for all $i\neq j$, with at least one off-diagonal pair present. Then for $b>0$ every off-diagonal term is positive, so
\[
\sum_{i,j=1}^{T}b^{2T-i-j}C_{ij}>R_0(b).
\tag{B.14}
\]
The same positive memory therefore costs strictly more.

\paragraph{Step 4: Profiling the gain leaves the demand measured against the memory cost.}
The variables $a$ and $c$ enter Eq.~\eqref{eq:appB-energy} only through their product $g=ca$, which we hold at its best value for each setting of the dial. Rewrite the energy as
\[
\begin{aligned}
E(g,b)=\frac12\sum_is_i^2
&-g\sum_{i=1}^{T}s_ib^{T-i}\\
&+\frac12g^2\sum_{i,j=1}^{T}b^{T-i}C_{ij}b^{T-j}.
\end{aligned}
\tag{B.15}
\]
Assume $C\succ0$. Then the regularization sum equals $v^\top Cv>0$ because the last entry of $v$ is one. The derivative with respect to $g$ vanishes at
\[
\begin{aligned}
\frac{\partial E}{\partial g}
&=-\sum_{i=1}^{T}s_ib^{T-i}
+g\sum_{i,j=1}^{T}b^{T-i}C_{ij}b^{T-j}=0,\\
g^\star(b)
&=\frac{\sum_{i=1}^{T}s_ib^{T-i}}
{\sum_{i,j=1}^{T}b^{T-i}C_{ij}b^{T-j}}.
\end{aligned}
\tag{B.16}
\label{eq:appB-profiled-gain}
\]
Substituting $g^\star$ back into the energy leaves the profiled energy
\[
E^\star(b)
=\frac12\left[
\sum_is_i^2-g^\star(b)\sum_{i=1}^{T}s_ib^{T-i}
\right].
\tag{B.17}
\label{eq:appB-profiled-objective}
\]
The target-size term is fixed, so minimizing $E^\star$ over $b$ maximizes the quantity subtracted from it, $g^\star(b)\sum_{i=1}^{T}s_ib^{T-i}$, the memory-weighted demand taken at the best gain the memory cost allows. With the gain held there, what is left for the dial is the data-driven term measured against the regularization term. This is the correlated-input version of the profiling step of \citet{proca2025learning}, in which the white cost of Eq.~\eqref{eq:appB-white-cost} stands where the covariance sum stands here.

\paragraph{Step 5: On the gain-optimal locus the dial runs downhill on the profiled energy.}
An interior fixed point has $a\neq0$, $c\neq0$, and $\dot a=\dot c=0$. Factoring the first and third equations of Eq.~\eqref{eq:appB-flow} gives
\begin{align*}
\tau\dot a&=c\Bigl(
\sum_{i}s_ib^{T-i}-g\sum_{i,j}b^{T-i}C_{ij}b^{T-j}
\Bigr),\\
\tau\dot c&=a\Bigl(
\sum_{i}s_ib^{T-i}-g\sum_{i,j}b^{T-i}C_{ij}b^{T-j}
\Bigr).
\end{align*}
Thus every interior fixed point lies on the gain-optimal locus $g=g^\star$ of Eq.~\eqref{eq:appB-profiled-gain}. Substituting this identity into the $b$-equation, with the derivatives of Eq.~\eqref{eq:appB-reward-cost}, yields
\[
\begin{aligned}
\left.\tau\dot b\right|_{g=g^\star}
&=g^\star\sum_{i=1}^{T-1}(T-i)s_ib^{T-i-1}\\
&\quad-\frac12(g^\star)^2\sum_{i,j=1}^{T}(2T-i-j)b^{2T-i-j-1}C_{ij},
\end{aligned}
\tag{B.18}
\]
which is $-\,\partial E/\partial b$ read at the best gain. Because the gain sits where $\partial E/\partial g$ vanishes, its own motion with the dial adds nothing to the derivative of the profiled energy,
\[
\frac{dE^\star}{db}
=\left.\frac{\partial E}{\partial b}\right|_{g=g^\star}
+\underbrace{\left.\frac{\partial E}{\partial g}\right|_{g=g^\star}}_{=0}
\frac{dg^\star}{db},
\tag{B.19}
\]
the second term vanishing by Eq.~\eqref{eq:appB-profiled-gain}, so
\[
\left.\tau\dot b\right|_{g=g^\star}=-\frac{dE^\star}{db}.
\tag{B.20}
\label{eq:appB-profiled-flow}
\]
On the gain-optimal locus the dial runs downhill on the profiled energy of Eq.~\eqref{eq:appB-profiled-objective}: it climbs whenever the demand sum's relative slope beats half the regularization sum's, the comparison Appendix~C reads at zero memory.

An interior fixed point therefore forces $dE^\star/db=0$. Conversely, any critical point of $E^\star$ at which the memory-weighted demand $\sum_is_ib^{T-i}$ is nonzero gives a nonzero profiled gain and hence an interior fixed point. A critical point at which that demand vanishes has $g^\star=0$ and is not interior.

The recurrent map is stable only for $|b|<1$, so the mode's domain has boundary $b\to\pm1$. A minimizer of $E^\star$ on that boundary need not be a fixed point, since $dE^\star/db$ need not vanish at a one-sided minimum. It can vanish there, however: for a flat demand $s_i\equiv1$ the two relative slopes coincide at $b\to1$ for any stationary $C$, so the drift vanishes at the boundary. The endpoint of interest below is $b=0$, where no such coincidence arises.

\paragraph{Why these simulations carry no sampling error.}
The main text states in a single clause that the flow reaches a dataset only
through two families of statistics. The long form is this.
Equation~\eqref{eq:appA-mode-energy} shows that the parameter-dependent part of
the mode-$\alpha$ loss depends on the trajectories only through the
memory-weighted demand $\sum_is_{\alpha,i}b_\alpha^{T-i}$, built from the
input-target profile $s_{\alpha,i}$ of Assumption~1, and the regularization sum
$\sum_{i,j}b_\alpha^{T-i}C_{ij}b_\alpha^{T-j}$, built from the temporal
covariance $C_{ij}$ of
Assumption~3. Those same two objects are all that enter the gradients in
Eq.~\eqref{eq:appB-flow}, and no further functional of the sample survives: the
one remaining term $\langle\widetilde y_\alpha^2\rangle_P$ does not depend on
the weights and drops out of every derivative. Fixing $s$ and $C$ therefore
fixes the loss surface exactly, and with it the entire trajectory of the flow.
Every run reported here substitutes population values for those two statistics
instead of averaging over drawn trajectories, so its curves are the exact
finite-$T$ population dynamics and carry no sampling error.

A finite training set of $P$ trajectories differs from this in one respect only.
It replaces $s$ and $C$ by the sample estimates
$\langle\widetilde y_\alpha\widetilde x_{\alpha,i}\rangle_P$ and
$\langle\widetilde x_{\alpha,i}\widetilde x_{\alpha,j}\rangle_P$ of
Eqs.~\eqref{eq:appA-task-statistic} and~\eqref{eq:supp-appA-input-statistic}.
The resulting dynamics are the exact population dynamics of that perturbed
pair, so all finite-$P$ error is the estimation error of two second moments, of
order $P^{-1/2}$, and none of it is a separate property of the flow. The
boundary of Lemma~\ref{lem:boundary} in particular is a ratio of two entries of $C$, so at
finite $P$ it moves by the sampling error in that one ratio and by nothing
else. This is also what lets Appendix~E pose each real series as a population
problem with its own estimated covariance: the estimate is the only place the
data enters.

\paragraph{Simulation details.}
The simulations in Figure~\ref{fig:dynamics-price}a, Figure~\ref{fig:grid-dynamics}, and the full-grid appendix figure integrate Eq.~\eqref{eq:appB-flow} by explicit Euler at $\tau=1$ with step $10^{-3}$ in training time, from the aligned small-weight start $a=c=0.05$, $b=0$. Figure~\ref{fig:dynamics-price}a runs to $t_\theta=40$, and the grids run to $t_\theta=30$. At $\lambda=0.6$ and $T=20$, the settled dials match the energy minima to three decimals: $b=0.600,0.408,0.000,-0.500$ at $\rho=0,0.3,0.6,0.9$, respectively. Figure~\ref{fig:recruit-grid} instead trains the full weight matrices of the model from random weights of scale $0.05$, twenty networks per cell, by explicit Euler with step $4\times10^{-3}$ to $t_\theta=400$, at $T=20$, $\lambda=0.6$, and $\rho\in\{0.7,0.75,0.8,0.85,0.9\}$, for widths one and two. The dials plotted there are the eigenvalues of the trained recurrent matrix. No run is stopped early; all are integrated to the horizon stated.

\begin{figure*}[tp]
\centering
\includegraphics[width=\textwidth]{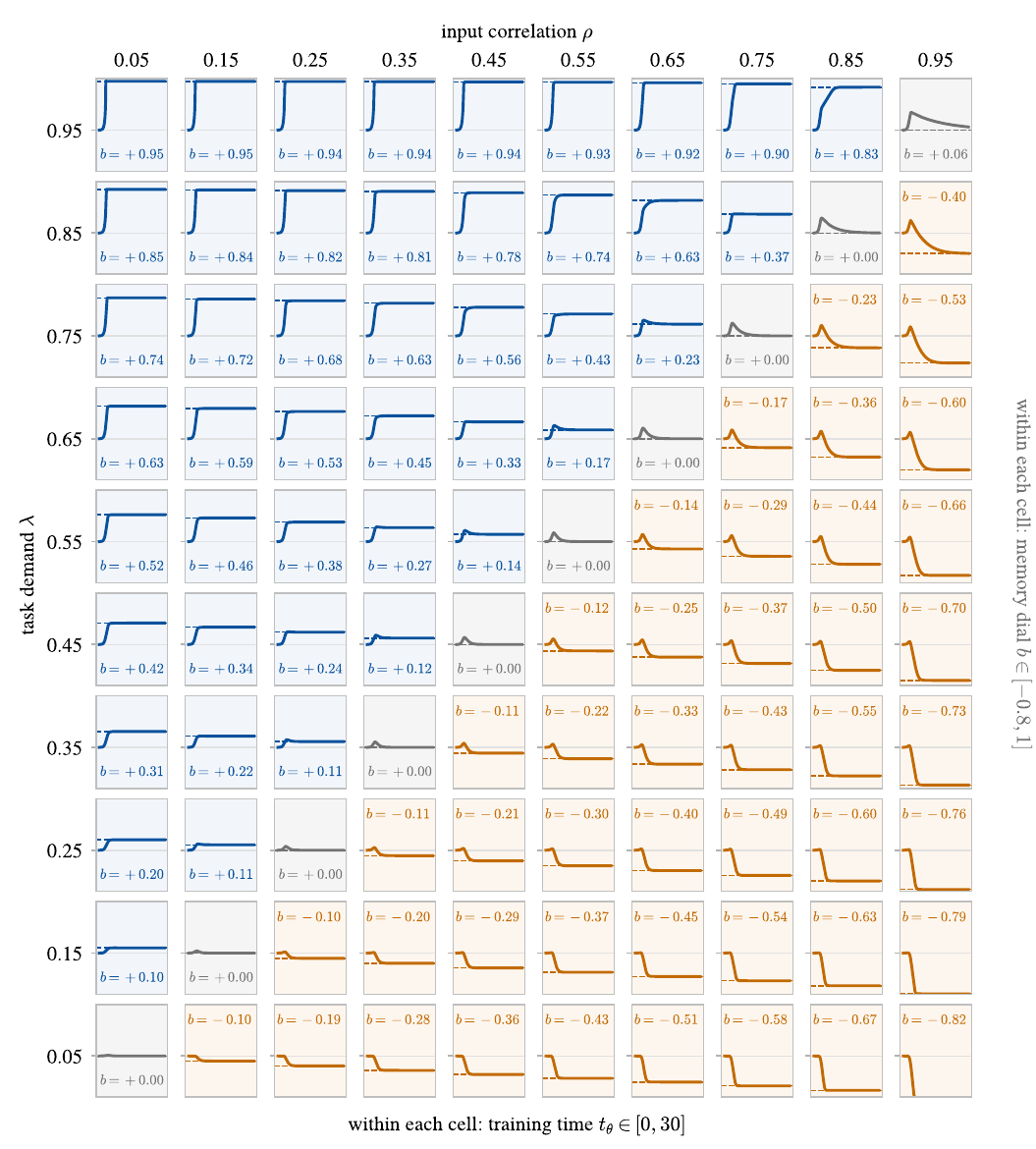}
\caption{The full grid behind Figure~\ref{fig:grid-dynamics}. Each
cell trains a network at one pairing of task demand $\lambda$
(rows, $0.95$ down to $0.05$) and input correlation $\rho$
(columns, $0.05$ to $0.95$), AR(1) inputs with the geometric task
at $T = 20$, extending the $4 \times 6$ grid of
Figure~\ref{fig:grid-dynamics} to $10 \times 10$. Solid curves are
the simulated trajectories of the memory dial $b$, dashed lines the
energy minima, and the printed value the settled $b$. The diagonal
$\lambda = \rho$ (gray) again separates memory ($\lambda > \rho$,
blue) from change detection ($\lambda < \rho$, orange). The corner
$\lambda = \rho = 0.95$ carries the slowest dynamics of the grid
and is still en route to its minimum at zero when the training
window ends.}
\label{fig:grid-full}
\end{figure*}

\paragraph{Numerical settings, what was tried and why these.}
Three numbers are free in every run, the Euler step, the horizon the run is integrated to, and the scale of the starting weights. Each was swept while the other two held at the reported value, and the setting kept is the cheapest one that meets the paper's own acceptance test, that the settled dial match the minimum of the energy \eqref{eq:energy}, restated as Eq.~\eqref{eq:appB-energy} here, to three decimals. For the single-mode flow the step was tried at $8\times10^{-3}$, $4\times10^{-3}$, $10^{-3}$ and $2.5\times10^{-4}$, and every value meets the test at $\rho=0$, $0.6$ and $0.9$, so the step is not what binds. The horizon is. At $t_\theta=15$ the boundary cell $\rho=\lambda$ has not finished settling and misses the test by $8.8\times10^{-4}$, while $t_\theta=30$ and $t_\theta=60$ both meet it, so the grids run to $30$. That test is met by every cell of the $4\times6$ grid; on the $10\times10$ grid of Figure~\ref{fig:grid-full} the two slowest diagonal cells are finite-time settles still en route to zero at this horizon. The starting scale was tried at $0.01$, $0.05$ and $0.2$ and moves no settled dial. For the full-matrix training of Figure~\ref{fig:recruit-grid} at $\rho=0.9$, the hardest cell, the step was tried at $8\times10^{-3}$, $4\times10^{-3}$ and $2\times10^{-3}$, the horizon at $200$, $400$ and $800$, and the starting scale at $0.01$, $0.05$, $0.2$ and $0.5$. Under every one of these settings the median settled dials over five networks split into the task demand $0.6$ and zero, both to three decimals. The recruitment result therefore does not rest on the small-weight start, and survives a fifty-fold change in the starting scale.

\paragraph{Software and hardware.}
Every simulation runs on the CPU in Python 3.12 with NumPy, SciPy and Matplotlib, with no GPU. The slowest figure, Figure~\ref{fig:recruit-grid}, trains two hundred networks in five minutes; the others are faster. Its random starts come from NumPy's default generator, seeded once per cell by $\lfloor 1000+10\rho+n\rfloor$ with $\rho$ the input correlation and $n$ the width, the truncation the script applies, so every run reported here is reproduced exactly by rerunning the script.
\section{C. Lag-One Transition and Tail Invariance}
\label{app:lag-one}

This appendix proves the second Finding and Lemma~\ref{lem:boundary}, the memory boundary. Throughout, $\lambda^\star$ denotes the critical value of the demand ratio $s_{T-1}/s_T$, which on the geometric task $s_i=\lambda^{T-i}$ is the rate $\lambda$ itself. Once the gain $g=ca$ is held at its best value for each setting of the dial, Appendix~B reduces learning to descent on the energy that remains, a function of the dial alone. Positive memory appears when the one-sided slope of that energy at the memoryless point $b=0$ turns negative, equivalently when the dial's drift there turns positive. Computing that slope yields
\[
\boxed{\lambda^\star=\frac{C_1}{C_0}},
\tag{C.1}
\label{eq:appC-boundary-preview}
\]
where $C_0=C_{TT}$ is the newest-input variance and $C_1=C_{T,T-1}$ is the lag-one covariance. For the geometric task, $s_{T-1}/s_T=\lambda$; for an AR(1) input, $C_1/C_0=\rho$. Hence the transition is $\lambda^\star=\rho$. The calculation is local and exact: it determines when the nonnegative-memory solution leaves the endpoint. The later steps explain why neither the sequence length nor covariance lags two and above can enter this boundary. Negative $b$ remains allowed by the unconstrained dynamics and corresponds to the change-detecting regime in Figure~\ref{fig:dynamics-price}a.

\paragraph{The energy at the best gain.}
The modes $a$ and $c$ enter the energy of Lemma~\ref{lem:energy} only through the gain $g=ca$. From Appendix~B, optimizing that gain at fixed $b$ gives
\[
\begin{aligned}
g^\star(b)&=\frac{\sum_{i=1}^{T}s_ib^{T-i}}
{\sum_{i,j=1}^{T}b^{T-i}C_{ij}b^{T-j}},\\
E^\star(b)&=\frac12\left[
\sum_{i=1}^{T}s_i^2-g^\star(b)\sum_{i=1}^{T}s_ib^{T-i}
\right].
\end{aligned}
\tag{C.2}
\label{eq:appC-profiled-objective}
\]
Two sums carry the whole dependence on the dial. The first, $\sum_is_ib^{T-i}$, is the demand the task places on the memory, the sum inside the data-driven term of Lemma~\ref{lem:energy}. The second, $\sum_{i,j}b^{T-i}C_{ij}b^{T-j}$, is that lemma's regularization sum, what the memory costs. The task profile $s_i$ comes from Assumption~1, and $C_{ij}$ is the temporal covariance from Assumptions~2--3.

We study the feedforward-versus-recurrent question on the nonnegative axis $b\geq0$. There $b=0$ is feedforward and memoryless, while $b>0$ is genuinely recurrent, so within this scope
\[
b^\star=\operatorname*{arg\,min}_{b\geq0}E^\star(b).
\tag{C.3}
\]
\citet{proca2025learning} restrict $b\in[0,1)$ for stability, but the endpoint analysis below sees only a right-neighborhood of zero, so the upper bound plays no role. Negative $b$ stores the past with alternating signs and is the separate change-detecting solution shown in Figure~\ref{fig:dynamics-price}a.

Profiling out the gain leaves $\tau\dot b=-(E^\star)'(b)$, gradient descent on $E^\star$ over training time. For $b>0$ the dial descends $E^\star$. At $b=0$, the projected nonnegative flow leaves the endpoint only when $E^\star$ slopes downward; when $(E^\star)'(0)\geq0$, projection holds the dial at zero.

The question is therefore local to the endpoint. Steps~1 and~2 show that at zero memory the regularization sum is the variance $C_0$, positive by Assumption~2, and the demand sum is $s_T$, positive whenever the task reads the newest input. Both sums therefore stay positive in a neighborhood of the origin, and their logarithmic slopes there are well defined. The best gain at the endpoint is then $g^\star(0)=s_T/C_0$, and the endpoint drift carries the positive prefactor $s_T^2/C_0$. Hence the sign of that drift is the sign of the demand slope $s_{T-1}/s_T$ minus the overlap slope $C_1/C_0$. The feedforward point is retained when the demand slope is the smaller and released when it is the larger. Equality is the transition. We compute the regularization sum's logarithmic slope in Step~1 and the demand sum's in Step~2, combine them in the endpoint drift in Step~3, read the boundary in Step~4, and study the release in Step~5. Steps~6 and~7 show that only lag one enters and that white input recovers the earlier cost.

\paragraph{Step 1: The regularization sum's logarithmic slope is $2C_1/C_0$.}
Start from
\[
\sum_{i,j=1}^{T}b^{T-i}C_{ij}b^{T-j}
=\sum_{i,j=1}^{T}b^{2T-i-j}C_{ij}.
\tag{C.4}
\]
At $b=0$, every monomial with $2T-i-j\geq1$ vanishes, leaving the $(T,T)$ term alone:
\[
\left.\sum_{i,j=1}^{T}b^{2T-i-j}C_{ij}\right|_{b=0}
=C_{TT}=C_0.
\tag{C.5}
\]
Differentiate term by term:
\[
\frac{d}{db}\sum_{i,j=1}^{T}b^{2T-i-j}C_{ij}
=\sum_{i,j=1}^{T}(2T-i-j)b^{2T-i-j-1}C_{ij}.
\tag{C.6}
\]
The $(T,T)$ term is absent because its prefactor is zero. At $b=0$, only the two pairs satisfying $2T-i-j=1$ survive, so
\begin{align*}
\left.\frac{d}{db}\sum_{i,j=1}^{T}b^{2T-i-j}C_{ij}\right|_{b=0}
&=C_{T,T-1}+C_{T-1,T}\\
&=2C_{T,T-1}=2C_1.
\end{align*}
Dividing that slope by the value $C_0$ the sum takes at zero gives its
logarithmic slope there, $2C_1/C_0$.
Half of this slope, the overlap $C_{T,T-1}$ measured against the variance $C_{TT}$, is the overlap slope $C_1/C_0$ that Step~3 sets against the demand.

\paragraph{Step 2: The demand sum's logarithmic slope is $s_{T-1}/s_T$.}
From
\[
\sum_{i=1}^{T}s_ib^{T-i}
=s_T+s_{T-1}b+\cdots+s_1b^{T-1},
\tag{C.7}
\]
we have
\[
\begin{aligned}
\left.\sum_{i=1}^{T}s_ib^{T-i}\right|_{b=0}&=s_T,\\
\frac{d}{db}\sum_{i=1}^{T}s_ib^{T-i}
&=\sum_{i=1}^{T-1}(T-i)\,s_ib^{T-i-1},\\
\left.\frac{d}{db}\sum_{i=1}^{T}s_ib^{T-i}\right|_{b=0}&=s_{T-1}.
\end{aligned}
\tag{C.8}
\]
Therefore
\[
\left.\frac{d}{db}\log\sum_{i=1}^{T}s_ib^{T-i}\right|_{b=0}
=\frac{s_{T-1}}{s_T}.
\tag{C.9}
\]
For the geometric task $s_i=\lambda^{T-i}$, this ratio is $\lambda$.

\paragraph{Step 3: The endpoint drift sets the demand slope against the overlap slope.}
On the gain-optimal locus, the dial's drift is the flow of Eq.~\eqref{eq:appB-profiled-flow}, which written out in the two sums reads
\[
\begin{aligned}
\left.\tau\dot b\right|_{g=g^\star}
&=g^\star(b)\sum_{i=1}^{T-1}(T-i)\,s_ib^{T-i-1}\\
&\quad-\frac12\bigl(g^\star(b)\bigr)^2
 \sum_{i,j=1}^{T}(2T-i-j)b^{2T-i-j-1}C_{ij}.
\end{aligned}
\tag{C.10}
\]
Evaluate it at the endpoint. Steps~1 and~2 supply the two values at zero memory, $C_0$ and $s_T$, hence the best gain $g^\star(0)=s_T/C_0$, and the two derivatives there, $2C_1$ and $s_{T-1}$, so
\begin{align*}
\left.\tau\dot b\right|_{b=0,\,g=g^\star}
&=g^\star(0)\bigl(s_{T-1}-g^\star(0)C_1\bigr)\\
&=\frac{s_T^2}{C_0}\left(\frac{s_{T-1}}{s_T}-\frac{C_1}{C_0}\right).
\end{align*}
The two logarithmic slopes enter exactly here, the factor of two on the regularization side cancelled by the $\tfrac12$ the squared gain carries. For the geometric task, with $\rho=C_1/C_0$,
\[
\left.\tau\dot b\right|_{b=0,\,g=g^\star}
=\frac{s_T^2}{C_0}\,(\lambda-\rho).
\tag{C.11}
\]
Thus memory pays at the margin only when the task reaches back faster than the input already predicts itself.

\paragraph{Step 4: The transition as a boundary minimizer and boundary equilibrium.}
Because $s_T>0$ and $C_0>0$, the prefactor $s_T^2/C_0$ is positive, so the sign of the endpoint drift is the sign of $s_{T-1}/s_T-C_1/C_0$. On $b\geq0$, the feedforward point is the left endpoint, so its optimality is a one-sided condition rather than an interior-stationarity condition.

\emph{Boundary-minimizer reading.}
The endpoint is a strict local minimizer of $E^\star$ when the endpoint drift is negative and is not a minimizer when it is positive. Equivalently,
\[
\left.\tau\dot b\right|_{b=0,\,g=g^\star}\leq0
\quad\Longleftrightarrow\quad
\frac{s_{T-1}}{s_T}\leq\frac{C_1}{C_0}.
\tag{C.12}
\]
When the inequality is strict, the projected flow points toward the feedforward endpoint; when it fails, the endpoint releases a positive recurrent branch. On the full signed axis, a demand slope below the overlap slope instead sends the optimum toward $b<0$, producing change detection. Thus the result locates the turn-off of positive recurrence; global optimality on the nonnegative axis is checked numerically in Step~5.

\emph{Boundary-equilibrium reading.}
The unconstrained endpoint drift is
\[
\begin{aligned}
\left.\tau\dot b\right|_{b=0^+,\,g=g^\star}
&=g^\star(0)\bigl(s_{T-1}-g^\star(0)C_1\bigr),\\
g^\star(0)&=\frac{s_T}{C_0}.
\end{aligned}
\tag{C.13}
\]
The sign test is read on the gain-optimal locus, as in Eq.~\eqref{eq:appB-profiled-flow}. From a small-weight start the gain $g=ca$ equilibrates onto $g^\star$ far faster than the dial moves, so the test applies after that transient rather than at $t_\theta=0$, where the raw drift $\tau\dot b=g(s_{T-1}-gC_1)$ is positive for any small $g$ whenever the task reads the second-newest input.
At equality the drift vanishes, so $b=0$ is a critical point of $E^\star$; with the curvature condition of Step~5 it is a local minimizer of $E^\star$ and hence a rest point of $\tau\dot b=-(E^\star)'$, which is the settling claimed on the boundary. When $s_{T-1}/s_T<C_1/C_0$, this drift is negative and the unconstrained dial moves toward change detection. Projected onto $b\geq0$, the negative drift is clipped to zero, so the endpoint is a stable equilibrium of the projected flow. When $s_{T-1}/s_T>C_1/C_0$, the drift is positive and recurrence switches on. Both readings leave zero at the same margin:
\[
\frac{s_{T-1}}{s_T}=\frac{C_1}{C_0}.
\tag{C.14}
\label{eq:appC-margin}
\]
On the geometric task,
\[
\lambda^\star=\frac{C_1}{C_0}=\rho.
\tag{C.15}
\]

\paragraph{Step 5: Local release and the global numerical check.}
Past the margin, an interior minimizer solves $(E^\star)'(b^\star)=0$. Expanding near the origin gives
\[
(E^\star)'(b)\approx(E^\star)'(0)+(E^\star)''(0)\,b,
\qquad
(E^\star)''(0)\neq0.
\tag{C.16}
\]
When $(E^\star)''(0)>0$,
\[
b^\star\approx-\frac{(E^\star)'(0)}{(E^\star)''(0)}\propto\lambda-\rho,
\tag{C.17}
\]
so the positive branch is released continuously. This curvature condition holds in the AR(1)/geometric example and in the continuous crossings reported in the main text; it is an additional local condition, not a consequence of the slope calculation alone.

The endpoint argument does not by itself exclude a distant interior minimizer that undercuts $b=0$ and produces a first-order jump. We therefore check global optimality numerically. For every covariance and sequence length tested in the AR(1)-like regime, the global recurrent-feedforward transition remains at $\lambda^\star=\rho$. Outside that regime this check does not hold: Appendix~E identifies the strongly periodic and highly persistent covariances on which a distant minimum overtakes the endpoint instead, and Figure~\ref{fig:grid-dynamics} traces the same boundary. The first-order transition reported by \citet{proca2025learning} for a separate construction with an added feedforward computation $\kappa$ is not the transition studied here.

\paragraph{Step 6: Neither lag two and above nor the sequence length enters the boundary.}
The boundary depends only on the regularization sum's value and slope at zero, the coefficients of $b^0$ and $b^1$. Each covariance entry $C_{ij}$ appears at power $b^{2T-i-j}$. For lag $d=|i-j|$,
\[
2T-i-j\geq|i-j|=d.
\tag{C.18}
\]
Hence lag $d$ first appears no earlier than order $b^d$. Collecting the first three orders,
\[
\sum_{i,j=1}^{T}b^{2T-i-j}C_{ij}
=\underbrace{C_0}_{b^0}
+\underbrace{2C_1}_{b^1}b
+\underbrace{(2C_2+C_0)}_{b^2}b^2
+O(b^3).
\tag{C.19}
\]
Every lag $d\geq2$ is invisible to those first two coefficients, so the transition depends on $C$ only through $C_1/C_0$. The sequence length is invisible for the same reason: the boundary reads $C$ only at $C_{TT}$ and $C_{T,T-1}$, which under Assumption~3 are $C_0$ and $C_1$ whatever $T$ is, and reads the task only through $s_{T-1}/s_T$, which on the geometric task is $\lambda$ whatever $T$ is. Lengthening the window adds only higher-order coefficients and leaves the margin untouched. Steps~1--3 touch $C$ only through the two entries $C_{TT}$ and $C_{T,T-1}$ and use only its symmetry, never the stationarity of Assumption~3. The boundary therefore locates correctly on covariances that are not stationary, which is the case on the estimated covariances of Appendix~E. For such a covariance, the same calculation gives $s_{T-1}/s_T=C_{T,T-1}/C_{TT}$. Tail invariance concerns only the transition location; the magnitude of $b^\star$ away from the transition still depends on the full covariance.

\paragraph{Step 7: The white limit recovers Proca et al.'s cost.}
Set $C=I$. Then $C_{ij}=\delta_{ij}$, $C_0=1$, and $C_1=0$, so
\[
\sum_{i,j=1}^{T}b^{2T-i-j}\delta_{ij}
=\sum_{i=1}^{T}b^{2T-2i}.
\tag{C.20}
\]
This is Proca et al.'s white cost. It contains only even powers, so
\[
\left.\frac{d}{db}\sum_{i=1}^{T}b^{2T-2i}\right|_{b=0}=0,
\qquad
\left.\frac{d}{db}\log\sum_{i=1}^{T}b^{2T-2i}\right|_{b=0}=0.
\tag{C.21}
\]
The overlap slope vanishes, so the endpoint drift reduces to the data-driven contribution alone:
\[
\left.\tau\dot b\right|_{b=0,\,g=g^\star}
=s_T\,s_{T-1}.
\tag{C.22}
\]
The boundary therefore returns to the uncorrelated value $\lambda^\star=0$: with no overlap to pay against, any demand at all for the second-newest input switches memory on.

Collecting Steps~1--4, the lag invariance of Step~6, and the white limit of Step~7 gives
\[
\boxed{\lambda^\star=\frac{C_1}{C_0}},
\tag{C.23}
\label{eq:appC-final-boundary}
\]
the recurrent-feedforward transition, exact and independent of $T$, invariant to every covariance lag $C_d$ with $d\geq2$, and equal to zero in the white-input case. Its coincidence with the global transition is established locally and confirmed numerically in Step~5 and Figure~\ref{fig:grid-dynamics}.

\paragraph{Machine verification.}
Lean 4, version 4.31.0 \citep{demoura2021lean4}, with its mathematical library mathlib \citep{mathlib2020}, machine-checks the identities of Lemmas~\ref{lem:energy} and~\ref{lem:boundary}, the per-mode energy and the memory boundary, and of this appendix: for the AR(1) covariance $C_{ij}=\rho^{|i-j|}$ with the demand sum left arbitrary, the regularization sum written as a quadratic form in the memory weights $b^{T-i}$, its reduction at $C=I$ to the white cost $\sum_ib^{2T-2i}$, the three gradient-flow equations, the balanced invariant $a^2-c^2$ recorded in Appendix~B, the reduced energy and its minimality, the fixed-point identity that on the gain-optimal locus the dial's drift is minus the slope of that reduced energy, and the two origin slopes, $2C_1/C_0$ for the regularization sum and the sign of the endpoint drift. The dynamics, fixed-point, and slope identities are proved for an arbitrary differentiable pair of demand and regularization sums; their concrete derivatives are checked numerically, and the $T\to\infty$ limit and the decoupling of Appendix~A remain verified numerically. The proofs depend only on the standard axioms.
\section{D. Optimality Gap and Feedthrough Capacity}
\label{app:optimality-gap}

This appendix supplies the representation argument behind the third Finding, that correlation demands a feedthrough term. Appendices~A--B show that one aligned hidden dimension per task mode produces a single exponential impulse response $gv(b)$. Here we compare that restricted family with the unconstrained per-mode optimum, the Wiener/GLS filter $w^{*}=C^{-1}s$. Because every single-exponential response is an admissible full response, the difference between the two optimum values is nonnegative. For the AR(1)/geometric example, $C^{-1}s$ separates into a geometric pole and a newest-step spike. A second hidden dimension with recurrent eigenvalue zero represents the spike and closes the gap up to the finite-horizon boundary term.

We index task modes by $\alpha$ and timesteps by $i,j$, with $i=T$ denoting the newest input. Throughout this appendix a \emph{mode} is a task direction $\alpha$, a \emph{channel} is one hidden dimension with its own recurrent eigenvalue, and a \emph{tap} is one entry of the impulse response $w$. The symbol $N_h$ is the hidden width, and $e_T=(0,\ldots,0,1)^\top$ is the newest-step impulse.

\paragraph{Matrix form of the restricted model.}
From Appendix~A, the aligned weights are those in Eq.~\eqref{eq:appA-aligned}. Assumptions~2--3 give $\mathbb E[x_ix_j^\top]=C_{ij}I$, while Assumption~1 gives $\Sigma^{YX_i}=US_iV^\top$. Dropping the target-only constant, the trajectory-averaged loss in the per-step maps $M_i=W_yW_h^{T-i}W_x$ is
\[
L
=-\sum_i\operatorname{tr}\!\left({\Sigma^{YX_i}}^\top M_i\right)
+\frac12\sum_{i,j}C_{ij}\operatorname{tr}\!\left(M_iM_j^\top\right)
+\mathrm{const}.
\tag{D.1}
\label{eq:appD-trace-loss}
\]
This is the trace form of the data-driven and regularization terms in Lemma~\ref{lem:energy}, the per-mode energy. We show that the loss splits mode by mode in Step~1, that gradient flow preserves the split in Step~2, that a single exponential leaves a nonnegative gap in Step~3, that correlation opens the gap in Step~4, that no single ray can close it in Step~5, and that one eigenvalue-zero dimension per mode closes it in Step~6. The covariance-family table then records what replaces the single spike beyond AR(1).

\paragraph{Step 1: The energy decouples across modes.}
Appendix~A proved this split from the squared error; we restate it in trace form because Step~3 compares the aligned ray against the released vector $w$.

At an aligned point,
\[
M_i
=U\operatorname{diag}\!\left(c_\alpha b_\alpha^{T-i}a_\alpha\right)V^\top.
\tag{D.2}
\]
Orthonormality of the task frame gives
\begin{align*}
\operatorname{tr}\!\left({\Sigma^{YX_i}}^\top M_i\right)
&=\sum_\alpha s_{\alpha,i}c_\alpha b_\alpha^{T-i}a_\alpha,\\
\operatorname{tr}\!\left(M_iM_j^\top\right)
&=\sum_\alpha c_\alpha^2a_\alpha^2
b_\alpha^{T-i}b_\alpha^{T-j}.
\end{align*}
Each trace runs over the single mode index $\alpha$, with no term linking $\alpha$ to $\alpha'\neq\alpha$. Substituting into Eq.~\eqref{eq:appD-trace-loss} gives
\[
\begin{aligned}
L
&=\mathrm{const}
+\sum_\alpha\Bigg[
-c_\alpha a_\alpha\sum_{i=1}^{T}s_{\alpha,i}b_\alpha^{T-i}\\
&\qquad\quad
+\frac12c_\alpha^2a_\alpha^2
\sum_{i,j=1}^{T}b_\alpha^{T-i}C_{ij}b_\alpha^{T-j}
\Bigg]\\
&=\mathrm{const}+\sum_\alpha E_{C,\alpha}.
\end{aligned}
\tag{D.3}
\]
Temporal correlation enters each summand only through the regularization sum $\sum_{i,j=1}^{T}b_\alpha^{T-i}C_{ij}b_\alpha^{T-j}$ and links no two modes, because $C_{ij}I$ is a scalar in feature space. (Verified: the full loss minus $\sum_\alpha E_{C,\alpha}$ vanishes to machine precision at random aligned points.)

\paragraph{Step 2: The aligned manifold is gradient-flow invariant.}
Differentiating Eq.~\eqref{eq:appD-trace-loss} in the linear map gives
\[
\frac{\partial L}{\partial M_i}
=-\Sigma^{YX_i}+\sum_jC_{ij}M_j.
\tag{D.4}
\]
At the aligned point,
\[
\begin{aligned}
\frac{\partial L}{\partial M_i}
&=U\operatorname{diag}(r_{\alpha,i})V^\top,\\
r_{\alpha,i}
&=-s_{\alpha,i}+c_\alpha a_\alpha\sum_jC_{ij}b_\alpha^{T-j}.
\end{aligned}
\tag{D.5}
\]
Applying the chain rule to $W_x$ and $W_h$ yields
\begin{align*}
\frac{\partial L}{\partial W_x}
&=V_h\operatorname{diag}\!\left(
\sum_i c_\alpha b_\alpha^{T-i}r_{\alpha,i}
\right)V^\top,\\
\frac{\partial L}{\partial W_h}
&=V_h\operatorname{diag}\!\left(
\sum_i(T-i)c_\alpha a_\alpha b_\alpha^{T-i-1}r_{\alpha,i}
\right)V_h^\top.
\end{align*}
The $W_y$ gradient is diagonal in the bases $U,V_h$ by the same contraction. No off-diagonal cross-mode component is created, so gradient flow from an aligned point never leaves the manifold. Coupling would require a spatially structured covariance misaligned with $V$, which the scalar feature-space covariance $C_{ij}I$ cannot produce. (Verified: 1500 full-matrix gradient-descent steps from an aligned start keep every off-diagonal entry and the gap to the scalar flow below $10^{-9}$.)

\paragraph{Step 3: A single exponential leaves a nonnegative gap.}
For a released impulse-response vector $w$, the per-mode loss is $\tfrac12w^\top Cw-s_\alpha^\top w$. Setting its gradient to zero gives the normal equations $\sum_jC_{ij}w_j=s_{\alpha,i}$ for $i=1,\dots,T$, which are Eq.~\eqref{eq:wiener} of the main paper. When $C\succ0$ they have the unique solution $w^{*}=C^{-1}s_\alpha$ \citep{kailath2000linear}. We use $w$ for the full vector of per-step taps and $a_\alpha$ for the scalar input connectivity.

An aligned mode maps a unit input at step $i$ to $c_\alpha b_\alpha^{T-i}a_\alpha$, so its impulse response lies on the ray $gv(b)$ with $g=c_\alpha a_\alpha$. As in the main text, $a_\alpha$ and $c_\alpha$ enter only through the gain $g$, which we hold at its best value for each setting of the dial. Restricting the per-mode loss to this ray and minimizing over $g$ and $b$ gives
\[
-\frac12\max_b
\frac{\bigl(s_\alpha^\top v(b)\bigr)^2}
{\sum_{i,j=1}^{T}b^{T-i}C_{ij}b^{T-j}}.
\tag{D.6}
\label{eq:appD-ray-value}
\]
At the best gain the objective is the data-driven term measured against the regularization term: the numerator is the square of the demand the ray meets, $s_\alpha^\top v(b)=\sum_{i=1}^{T}s_{\alpha,i}b^{T-i}$, and the denominator is the price the ray pays.
Releasing the ray gives the value
\[
-\frac12s_\alpha^\top C^{-1}s_\alpha.
\tag{D.7}
\label{eq:appD-wiener-value}
\]
Because the ray is a subset of all readouts, the aligned value can only exceed the unconstrained value. Summed over modes, their difference is the nonnegative optimality gap
\[
\begin{aligned}
\Delta(\rho)
&=\frac12\sum_\alpha\Bigg[
s_\alpha^\top C^{-1}s_\alpha\\
&\quad
-\max_b\frac{\bigl(s_\alpha^\top v(b)\bigr)^2}
{\sum_{i,j=1}^{T}b^{T-i}C_{ij}b^{T-j}}
\Bigg]\geq0.
\end{aligned}
\tag{D.8}
\label{eq:appD-gap}
\]
It vanishes for a mode exactly when $C^{-1}s_\alpha$ is proportional to some $v(b)$, that is, when the unconstrained optimum is itself a single exponential.

\paragraph{Step 4: Correlation opens the gap in the running example.}
For $C_{ij}=\rho^{|i-j|}$ and $s_{\alpha,i}=\sigma_\alpha\lambda_\alpha^{T-i}$, the full optimum has a newest-step tap and a geometric tail:
\begin{align*}
(C^{-1}s_\alpha)_T&\propto1,\\
(C^{-1}s_\alpha)_i&\propto(\lambda_\alpha-\rho)\lambda_\alpha^{T-i-1},
\qquad 2\leq i\leq T-1.
\end{align*}
The oldest step is the exception. The first row of the AR(1) precision matrix
carries diagonal entry $1$ rather than $1+\rho^2$ and only one neighbour, so
$(C^{-1}s_\alpha)_1$ exceeds the value the second line would give by the factor
$1/(1-\rho\lambda_\alpha)$. Equivalently, away from that edge,
\[
C^{-1}s_\alpha
\propto
\frac{\lambda_\alpha-\rho}{\lambda_\alpha}v(\lambda_\alpha)
+\frac{\rho}{\lambda_\alpha}e_T.
\tag{D.9}
\label{eq:appD-pole-spike-split}
\]
The ratio of the step-$(T-1)$ tap to the step-$T$ tap is $\lambda_\alpha-\rho$, matching the boundary margin in Lemma~\ref{lem:boundary}. The response lies on a single ray only when one coefficient vanishes: at $\rho=0$ it is the pure pole, and at $\rho=\lambda_\alpha$ it is the pure newest-step spike. Otherwise, the single-pole model cannot represent both components.

The exact gains follow from the tridiagonal inverse of the AR(1) covariance \citep{kailath2000linear}. For one mode, writing $\lambda=\lambda_\alpha$ and $\sigma=\sigma_\alpha$,
\begin{align*}
w_1^{*}
&=\sigma G\lambda^{T-1}
+\sigma\frac{\rho(\lambda-\rho)}{1-\rho^2}\lambda^{T-1},\\
w_i^{*}
&=\sigma G\lambda^{T-i},
\qquad 2\leq i\leq T-1,\\
w_T^{*}
&=\sigma\frac{1-\rho\lambda}{1-\rho^2}
 =\sigma(G+G_{\mathrm{spike}}),
\end{align*}
where
\[
G=\frac{(1-\rho\lambda)(1-\rho/\lambda)}{1-\rho^2},
\qquad
G_{\mathrm{spike}}=\frac{\rho(1-\rho\lambda)}{\lambda(1-\rho^2)}.
\tag{D.10}
\]
Since $1-\rho\lambda>0$, the tail gain has the sign of $1-\rho/\lambda$: it is positive below the diagonal, vanishes at $\rho=\lambda$, and is negative above it. The spike gain is nonnegative, zero only at $\rho=0$, and strictly increasing in $\rho$. The oldest entry contains an additional horizon term of order $\lambda^{T-1}$.

\paragraph{Step 5: No single ray carries a tail plus a spike.}
A ray $gv(\beta)$ has taps $g\beta^{T-i}$, so the ratio of consecutive taps is the constant $\beta$. At every interior step $2\leq i\leq T-2$, the optimum has ratio $w_{i}^{*}/w_{i+1}^{*}=\lambda$, so whenever the tail gain is nonzero, a matching ray must have $\beta=\lambda$. At the newest step,
\[
\frac{w_{T-1}^{*}}{w_T^{*}}=\lambda-\rho,
\tag{D.11}
\]
which differs from $\lambda$ whenever $\rho\neq0$. Therefore, for $0<\rho\neq\lambda$ and $T\geq4$, no ray at any rate equals the optimum. The exceptions are the pure pole at $\rho=0$ and the pure spike $v(0)$ at $\rho=\lambda$.

\paragraph{Step 6: One zero-eigenvalue dimension per mode closes the gap.}
A hidden dimension with recurrent eigenvalue $\beta$ has impulse response $g\beta^{T-i}$. Eigenvalue $\lambda_\alpha$ gives the pole $v(\lambda_\alpha)$. Eigenvalue zero gives $0^{T-i}=\delta_{iT}=e_T$, an instantaneous feedthrough at the newest step and nowhere else.

Setting the pole coefficient to $(\lambda_\alpha-\rho)/\lambda_\alpha$ and the feedthrough coefficient to $\rho/\lambda_\alpha$ reproduces Eq.~\eqref{eq:appD-pole-spike-split}. Two hidden dimensions per mode therefore represent the stationary large-$T$ optimum. The pole holds the tail $(\lambda_\alpha-\rho)\lambda_\alpha^{T-i-1}$ and the zero dimension holds the feedthrough spike $\delta_{iT}$.

The finite-window optimum departs from this shape at the oldest tap because the AR(1) inverse has a horizon boundary term \citep{kailath2000linear}. Two dimensions therefore match the optimum up to that term. In the AR(1)/geometric example, the demanded width doubles from $N$ to $2N$. For a general covariance, the optimum need not be pole-plus-spike, so the statement stays qualitative: correlation raises the width a task demands.

Trained full networks confirm the representation argument.
\texttt{experiments/capacity\_full.py} trains the full weight matrices, giving
the network no explicit feedthrough term, on $N=2$ task modes at rates
$\lambda=0.6$ and $0.35$ against an AR(1) input at $\rho=0.9$ and $T=20$,
twenty runs per width. At $N_h=N$ the trained objective settles at $-1.3176$,
which is the best single-pole value \eqref{eq:appD-ray-value} to fourteen
decimals: the width-$N$ network reaches the single-ray ceiling and stops there,
leaving the whole representation gap $\Delta=0.9596$ of
Eq.~\eqref{eq:appD-gap}. At $N_h=2N$ it reaches $-2.2772$, the Wiener value
\eqref{eq:appD-wiener-value}, leaving an optimality gap of
$3.3\times10^{-9}$. The learned spectrum of $W_h$ at that width is
$\{0.600,\,0.350,\,0.000,\,0.000\}$, one pole per task rate and one
eigenvalue-zero feedthrough channel per mode, which is the construction of
Step~6 built by gradient descent alone; the per-mode filters correlate with
$C^{-1}s_\alpha$ at $r>0.9999$. Raising the width to $N_h=3N$ leaves the
objective at $-2.2772$ and the gap at $3.7\times10^{-8}$, so the extra width
buys nothing. The residual gap at $2N$ is the finite-horizon edge term
discussed next, not a representation error: it is of order $\lambda^{2T}$,
which is $1.3\times10^{-9}$ here, and sits eight orders of magnitude below the
$0.9596$ that one missing dimension costs.

A single-mode version of the same test, \texttt{experiments/recruit\_full.py},
asks whether recruitment survives width the construction does not need. At
$T=20$, $\lambda=0.6$ and $\rho=0.9$ with one task mode, width one is held at
the single-ray gap of $2.85\times10^{-1}$ in twenty of twenty runs; width two
closes it to $3.29\times10^{-9}$ in twenty of twenty; and width four closes it
in twenty of twenty as well, with gaps from $3.1\times10^{-9}$ to
$1.3\times10^{-7}$. Extra width therefore neither blocks the recruitment nor
improves on it, which is the answer to the natural question of whether the
width-two result is an artifact of giving the network exactly the two
dimensions the construction asks for.

\paragraph{Beyond AR(1): covariance families.}
The AR(1) example with a geometric task is the simplest case of a broader
rule. For every covariance family in Table~\ref{tab:covariance-families}, we
form $w^{*}=C^{-1}s$ and search over feedthrough taps, recurrent poles, and
real damped-cosine channels. The selected components close the single-ray gap
at $T=20$ and $\lambda=0.6$. In every finite-order family, the interior of
$w^{*}$ retains the geometric tail at the task rate $\lambda$. The
covariance determines the correction around that tail: its whitening filter,
truncated to the observation window.

\begin{table*}[t]
\centering
\footnotesize
\begin{tabular}{@{}>{\raggedright\arraybackslash}p{0.17\textwidth}p{0.35\textwidth}p{0.40\textwidth}@{}}
\toprule
Input covariance & Shape of the optimum $w^{*} = C^{-1}s$ & Beyond a single ray, zero error demands \\
\midrule
AR(1), $C_{ij} = \rho^{\lvert i-j\rvert}$ & the $\lambda$-tail at gain $(1-\rho\lambda)(1-\rho/\lambda)/(1-\rho^{2})$, plus a newest-step spike & one feedthrough tap, the $D$ term \\
AR(2) & the $\lambda$-tail plus two newest-step taps with a sign flip & two taps, a width-two causal convolution \\
Seasonal AR, period $S$ & a positive season head on a negative $\lambda$-tail & $S$ taps, one full season; fewer buy almost nothing, and extra poles are worthless \\
MA(1), root $\theta$ & the $\lambda$-tail plus an alternating ray & one recurrent channel at $-\theta$; taps stall at rate $\theta^{2}$ per tap \\
ARMA(1,1) & the $\lambda$-tail plus an alternating ray & one channel at $-\theta$; the autoregressive tap is absorbed by the pole \\
Sum of two AR(1)s & the $\lambda$-tail, a spike, and a ray at the spectral zero $\theta_{*}$ & one tap plus one recurrent channel at $\theta_{*}$, which is neither input timescale \\
Damped cosine & the $\lambda$-tail plus a ray at the real moving-average root & one channel at that root plus one tap; no oscillatory channel is ever recruited \\
Exchangeable & the $\lambda$-tail on a flat negative pedestal & an integrator channel $b \to 1$; the gap follows $2.16\,(1-b)^{2}$, its infimum unattained \\
Long memory, ARFIMA($d$) & the $\lambda$-tail on a power-law tail & no finite menu; about $T/2$ channels piling up at $1$ \\
\bottomrule
\end{tabular}
\caption{The component beyond a single ray that zero error
demands, one covariance family per row, at $T = 20$,
$\lambda = 0.6$ with representative parameters. In every family
the interior of $w^{*}$ is the same geometric tail at the task's
own rate $\lambda$; the covariance decides what must be added
around it, and the addition is the input's whitening filter
truncated at the window. Writing the input's whitening filter, the reciprocal of the causal
spectral factor, as $a(z)/c(z)$, autoregressive order $p$ and moving-average
order $q$, the optimal transfer function is
$W(z) \propto a(z)/\bigl(c(z)\,(1-\lambda z)\bigr)$:
$\max(0,\, p-q)$ feedthrough taps, one recurrent channel per
moving-average zero, and a full season of taps per seasonal
factor. The recruited poles are never the input's own timescales,
only the task rate $\lambda$ and the zeros of the spectral
factorization. With the listed component in place the single-ray
gap, between $0.03$ and $2.2$ across rows, closes below $10^{-6}$
in every finite-order family; the last two rows are the limit
cases in which no finite menu attains the optimum.}
\label{tab:covariance-families}
\end{table*}

Every finite-order family also has an anti-causal edge term at the oldest
tap. Its size is of order $(\text{rate})^T$, and no causal channel spans it.
Its tap size runs from about $5\times10^{-6}$ to about $1\times10^{-3}$ across the reported configurations, while the residual loss it leaves is quadratic in that size and so stays below $10^{-6}$. The term vanishes
geometrically with $T$. Recomputing the released optima reproduces the
tabulated channel budgets to the printed precision.

Steps~1 and~2 keep the modes decoupled under gradient flow, Step~3 measures the loss left by a single exponential, Step~4 shows that correlation opens the gap, Step~5 proves that no single ray closes it, and Step~6 repairs it with an eigenvalue-zero dimension. Temporal correlation therefore cannot couple the task modes, but it can break the optimality of the aligned single-exponential form.

\paragraph{Code and data availability.}
The supplementary material accompanying this paper contains the verification code, the Lean proofs, the figure generators, the covariance-family scripts, the real-data pipeline, and one data file, \texttt{real\_data/series\_record.csv}, the recorded per-series outcome for the series whose sources need an account. It carries a \texttt{RUNNING.txt} covering layout, runtimes, the seeding rule, and how to check a run in one command. The series themselves are downloaded from the public sources cited in Appendix~E and are not redistributed here.
\section{E. Real-Data Validation}
\label{app:real-data}

This appendix backs the real-data claim of the second Finding with the task construction, the protocol, the nineteen-series table, a synthetic invariance check confirming both invariances numerically, the scope condition for Lemma~\ref{lem:boundary}, and a check of the negative-threshold regime.

\paragraph{Task construction.}
Each series supplies the inputs, and a linear teacher supplies the demand. Given the estimated covariance $C$ of the windowed series, targets are produced by
\[
w_{\mathrm{teach}}=C^{-1}s,
\qquad
s_i=\lambda^{T-i}.
\tag{E.1}
\]
The measured demand is then
\[
\frac1P\sum_{p=1}^Py_px_{p,i}
=(Cw_{\mathrm{teach}})_i
=s_i,
\tag{E.2}
\]
so the series' temporal structure lies entirely in $C$, while the input-target profile is exactly geometric at rate $\lambda$. Each series therefore poses the single-mode problem of Appendix~B with its own estimated covariance.

\paragraph{Protocol.}
A series is first differenced when it carries a trend, standardized to zero mean and unit variance, and cut into overlapping windows of length $T=20$ at stride one. Each series enters at its natural resolution: daily for weather, financial, and air-quality records over 2015--2024; full monthly history for the macroeconomic, sunspot, CO$_2$, and Ni\~no records; and 128-Hz samples for the EEG channels. Yield changes are monthly, the GloFAS discharges are weekly means, and solar radiation and wave height are hourly so their periodicities lie inside the window. The covariance $C$ is the second moment of the window matrix.

Six rows of the table are derived series rather than a published record, each built by one of three transforms, and
each is built one way only. A volatility series is the absolute log return of
the daily record over 2015--2024. A growth rate is the log difference of the
monthly record over its full history. The yield series is the change in the
monthly average of the daily rate. The two negatively correlated series below
are plain daily log returns over the same decade. With these transforms
and the resolutions above, every row of Table~\ref{tab:real-series} is rebuilt
from its source by \texttt{real\_data/build\_table.py}, whose manifest names the
source, the identifier, and the transform of each series.

The Mississippi appears twice, from two different providers, and the two are
kept apart throughout because they behave differently. The row in
Table~\ref{tab:real-series} is the GloFAS reanalysis discharge, read at weekly
means like the Rhine beside it, and its crossing is continuous. The
scope-condition series at the end of this appendix instead uses the USGS stream
gauge at Vicksburg, site 07289000, read as the raw daily record; that series is
far more persistent, at a measured threshold of $0.9986$ against the GloFAS
series' $0.8892$, and its crossing is first-order. Every mention of the
Mississippi in this appendix names its provider for that reason.

The measured threshold is the right-hand side of Eq.~\eqref{eq:boundary}, the correlation of the two most recent steps of $C$. The learned crossing is found by sixteen iterations of bisection in the demand rate $\lambda$, using the minimum of the energy \eqref{eq:energy} of Lemma~\ref{lem:energy} with the gain optimized. The bisection carries a systematic bias of approximately $+2\times10^{-4}$. It is not a grid effect: sixteen halvings leave the midpoint within $7.5\times10^{-6}$ of the root, and running more iterations does not shrink the offset. It comes from the tolerance that declares the settled dial nonzero, divided by the local slope of the settled dial in the demand rate, and its bracket must contain the crossing: the ceiling is $0.99$, which covers every series of Table~\ref{tab:real-series}, and is raised for the more persistent series of Table~\ref{tab:scope-series}.

Transitions are classified with a tight probe at the crossing plus and minus $3\times10^{-4}$, and a jump in the settled dial between the two marks a first-order transition. The probe settles the dial the same way the bisection does, at the global minimum of the energy. This is the only self-consistent choice, and it matters: settling by finite-time gradient flow from $b=0$ instead cannot detect a first-order transition at all, because a first-order crossing is exactly the case in which the global minimum moves to a distant dial that the local flow never reaches, so the flow returns the same near-zero value on both sides and reports every such series as continuous. Loose probes of $\pm0.01$ are also not used, because they can misclassify steep but continuous crossings.

The financial and macroeconomic series come from FRED \citep{fred2026};
weather variables from ERA5 \citep{hersbach2020era5}; reanalysis river
discharge from GloFAS \citep{harrigan2020glofas}, which supplies the Rhine, the
GloFAS Mississippi, and the Danube; PM$_{2.5}$ from CAMS
\citep{inness2019cams}; sunspots from SILSO \citep{silso2026}; EEG channels
from the eye-state corpus \citep{roesler2013eeg}; CO$_2$ from NOAA GML
\citep{noaagml2026}; Ni\~no~3.4 from ERSSTv5 \citep{huang2017ersst}; and the
Vicksburg stream gauge, site 07289000, from USGS \citep{usgs2026}. Every statistic below is computed from the same downloaded data, so the comparison between the learned crossing and the measured threshold is self-contained.

\paragraph{The nineteen-series table.}
Table~\ref{tab:real-series} lists the outcome per series, ordered by measured threshold. The thresholds span $0.1438$ to $0.9828$, and every learned crossing lies within $1.5\times10^{-4}$ of its measured threshold, inside the bisection bias. Every row is regenerated by \texttt{real\_data/build\_table.py}. Seven of the nineteen come from sources that serve plain CSV to anyone and are fetched on demand; the other twelve need a free Copernicus account or a manual archive download, and for those the archive ships \texttt{real\_data/series\_record.csv}, the recorded threshold, lag-one value, crossing, gap, and transition class of each, so that every cell of the table can be checked against a committed number without an account.

\begin{table}[t]
\centering
\footnotesize
\setlength{\tabcolsep}{3.5pt}
\begin{tabular}{@{}lrrrr@{}}
\toprule
Series & Thresh. & Lag-1 & Crossing & Gap \\
 & & & & \multicolumn{1}{r@{}}{\scriptsize$(10^{-5})$} \\
\midrule
FX volatility & $0.1438$ & $0.1433$ & $0.1439$ & $+12$ \\
precipitation & $0.2209$ & $0.2195$ & $0.2210$ & $+12$ \\
10-year-yield changes & $0.3077$ & $0.3079$ & $0.3077$ & $+5$ \\
S\&P volatility & $0.3937$ & $0.3961$ & $0.3938$ & $+15$ \\
gas volatility & $0.4587$ & $0.4602$ & $0.4588$ & $+9$ \\
US industrial prod.\ growth & $0.4982$ & $0.4897$ & $0.4983$ & $+7$ \\
cloud cover & $0.5079$ & $0.5082$ & $0.5079$ & $+8$ \\
Berlin wind & $0.5176$ & $0.5179$ & $0.5177$ & $+8$ \\
coastal wind & $0.5889$ & $0.5910$ & $0.5890$ & $+6$ \\
US inflation & $0.6152$ & $0.5941$ & $0.6153$ & $+8$ \\
Singapore wind & $0.6212$ & $0.6225$ & $0.6213$ & $+8$ \\
Melbourne min.\ temperature & $0.7100$ & $0.7103$ & $0.7100$ & $+6$ \\
Singapore temperature & $0.7437$ & $0.7447$ & $0.7438$ & $+7$ \\
PM2.5 & $0.7669$ & $0.7641$ & $0.7669$ & $+3$ \\
the Rhine, GloFAS & $0.7806$ & $0.7776$ & $0.7807$ & $+4$ \\
the Mississippi, GloFAS & $0.8892$ & $0.8848$ & $0.8892$ & $+2$ \\
sunspots & $0.9190$ & $0.9186$ & $0.9190$ & $+2$ \\
EEG O1 & $0.9673$ & $0.9673$ & $0.9673$ & $+1$ \\
EEG AF3 & $0.9828$ & $0.9828$ & $0.9828$ & $+0$ \\
\bottomrule
\end{tabular}
\caption{The nineteen series. Threshold is the measured boundary from Lemma~\ref{lem:boundary}, the correlation of the two newest steps of the estimated covariance; Lag-1 is the ordinary lag-one autocorrelation, the lag-one covariance averaged across the $T-1$ window positions and measured against the average variance; Crossing is the learned crossing found by bisection; and Gap is Crossing minus Threshold in units of $10^{-5}$. Every gap is at most $1.5\times10^{-4}$, inside the bisection bias, and every crossing is continuous under the tight probe. The bias is the $+2\times10^{-4}$ of the protocol above, and every gap in the table sits inside it, so no row is evidence of a real offset between the crossing and the threshold. Regenerated by \texttt{real\_data/build\_table.py}.}
\label{tab:real-series}
\end{table}

The ordinary lag-one column shows why the newest-step threshold, not a position average of the lag-one entries, is the correct statistic. For example, US inflation has lag-one value $0.5941$ but newest-step threshold $0.6152$, and the crossing follows the latter as the nonstationary form of Appendix~C, Step~6, predicts.

\paragraph{Tail invariance on synthetic covariances.}
Table~\ref{tab:synthetic-invariance} runs the same settle-and-bisect pipeline on synthetic covariances that share threshold $0.6$ while varying the sequence length and the covariance tail beyond the most recent pair. The crossing moves by at most $6.0\times10^{-5}$. This table needs no download, so it is reproducible in full today: \texttt{real\_data/invariance.py} builds all five covariances and prints the rows.

\begin{table}[t]
\centering
\footnotesize
\setlength{\tabcolsep}{3.5pt}
\begin{tabular}{@{}lrrr@{}}
\toprule
Covariance & Predicted & Learned & Gap $(10^{-5})$ \\
\midrule
AR(1) baseline, $T = 20$ & $0.6000$ & $0.6001$ & $+6$ \\
shorter window, $T = 12$ & $0.6000$ & $0.6001$ & $+6$ \\
longer window, $T = 30$ & $0.6000$ & $0.6001$ & $+6$ \\
AR(2) heavier tail, $\phi_2 = +0.2$ & $0.6000$ & $0.6001$ & $+9$ \\
AR(2) lighter tail, $\phi_2 = -0.3$ & $0.6000$ & $0.6000$ & $+3$ \\
\bottomrule
\end{tabular}
\caption{Tail invariance on synthetic covariances. Every row shares the ratio $C_{T-1,T}/C_{T,T}=0.6$ while the sequence length and the covariance tail vary. The AR(2) rows pin the lag-one correlation by the Yule--Walker relation $\phi_1=0.6(1-\phi_2)$. The learned crossing does not move beyond the bisection bias. Regenerated by \texttt{real\_data/invariance.py}, which needs no downloaded data.}
\label{tab:synthetic-invariance}
\end{table}

\paragraph{Negative thresholds.}
The two newest samples are negatively correlated in two series. The measured
threshold is $-0.154$ for daily S\&P returns and $-0.041$ for daily Bitcoin
returns. Equation~\eqref{eq:boundary} is therefore satisfied by every demand rate
$\lambda>0$, so no transition is predicted inside $(0,1)$. None is observed:
at $\lambda=0.01$, the settled dials are already positive, at $b=0.14$ and
$b=0.05$, respectively.

\paragraph{Scope condition.}
Lemma~\ref{lem:boundary} locates the release of the memoryless point exactly. On AR(1)-like covariances, that release is the complete transition and is continuous under the curvature condition in Appendix~C, Step~5. Strong periodicity inside the window or extreme persistence can instead produce a first-order transition: a distant energy minimum overtakes the endpoint, so the global crossing arrives below the measured threshold and the settled dial jumps at the crossing.

Table~\ref{tab:scope-series} lists the six series on which this happens. In
each the crossing arrives below the measured threshold, by between
$2\times10^{-3}$ and $1.5\times10^{-1}$, and the tight probe finds the settled
dial jumping from change detection to strong positive memory across it, by
between $1.2$ and $1.6$, far above the $0.05$ that separates the two classes.
The two rows that regenerate without credentials show the pattern: seasonal
CO$_2$ jumps from $-0.371$ to $+0.882$, and the raw daily Mississippi from
$-0.667$ to $+0.878$. Monthly unemployment changes, humidity, VIX, and surface pressure
instead remain on the continuous side, their settled dials moving by less than
$0.007$ under the same probe. The location claim of Lemma~\ref{lem:boundary} remains exact as a
local release condition; these structures affect only whether that release is
also the global, continuous transition.

\begin{table}[t]
\centering
\footnotesize
\setlength{\tabcolsep}{3.5pt}
\begin{tabular}{@{}llrrr@{}}
\toprule
Series & Resolution & Thresh. & Crossing & Shortfall \\
\midrule
seasonal CO$_2$ & monthly & $0.698$ & $0.550$ & $0.148$ \\
solar radiation & hourly & $0.943$ & $0.863$ & $0.080$ \\
Ni\~no~3.4 & monthly & $0.951$ & $0.935$ & $0.016$ \\
the Danube & weekly & $0.969$ & $0.964$ & $0.005$ \\
ocean wave height & hourly & $0.997$ & $0.994$ & $0.003$ \\
the Mississippi, USGS & daily, raw & $0.9986$ & $0.9964$ & $0.002$ \\
\bottomrule
\end{tabular}
\caption{The six scope-condition series, those whose crossing is
first-order. Thresh.\ is the measured boundary of Lemma~\ref{lem:boundary} as in
Table~\ref{tab:real-series}; Crossing is the learned crossing;
Shortfall is Threshold minus Crossing, the amount by which a distant
energy minimum overtakes the endpoint before the local release
condition fires. Every shortfall here is one to three orders of magnitude
above the bisection bias, unlike the gaps in
Table~\ref{tab:real-series}. The Mississippi row is the USGS stream
gauge at Vicksburg read as a raw daily record, a different provider
and a different resolution from the GloFAS Mississippi discharge
that enters Table~\ref{tab:real-series} at weekly means. The
CO$_2$ and Mississippi rows regenerate from public sources with no
credentials through \texttt{real\_data/build\_table.py}; the other
four need an account, and their recorded outcomes ship in
\texttt{real\_data/series\_record.csv} with the rest.}
\label{tab:scope-series}
\end{table}

\bibliography{aaai2027}

\begin{thebibliography}{32}
\providecommand{\natexlab}[1]{#1}

\bibitem[{Atanasov, Bordelon, and Pehlevan(2022)}]{atanasovneural}
Atanasov, A.; Bordelon, B.; and Pehlevan, C. 2022.
\newblock Neural Networks as Kernel Learners: The Silent Alignment Effect.
\newblock In \emph{International Conference on Learning Representations}.

\bibitem[{Atick and Redlich(1992)}]{6795763}
Atick, J.~J.; and Redlich, A.~N. 1992.
\newblock What Does the Retina Know about Natural Scenes?
\newblock \emph{Neural Computation}, 4(2): 196--210.

\bibitem[{Bordelon et~al.(2025)Bordelon, Cotler, Pehlevan, and
  Zavatone-Veth}]{bordelon2025dynamics}
Bordelon, B.; Cotler, J.; Pehlevan, C.; and Zavatone-Veth, J.~A. 2025.
\newblock Dynamics of Learning to Integrate in Linear Recurrent Neural
  Networks.
\newblock \emph{arXiv preprint arXiv:2503.18754}.

\bibitem[{Dambre et~al.(2012)Dambre, Verstraeten, Schrauwen, and
  Massar}]{Dambre2012}
Dambre, J.; Verstraeten, D.; Schrauwen, B.; and Massar, S. 2012.
\newblock Information Processing Capacity of Dynamical Systems.
\newblock \emph{Scientific Reports}, 2(1).

\bibitem[{de~Moura and Ullrich(2021)}]{demoura2021lean4}
de~Moura, L.; and Ullrich, S. 2021.
\newblock The {L}ean 4 Theorem Prover and Programming Language.
\newblock In \emph{Automated Deduction -- CADE 28}, volume 12699 of
  \emph{Lecture Notes in Computer Science}, 625--635. Springer.
\newblock Version 4.31.0, commit 68218e876d2a.

\bibitem[{{Federal Reserve Bank of St.\ Louis}(2026)}]{fred2026}
{Federal Reserve Bank of St.\ Louis}. 2026.
\newblock {FRED}, Federal Reserve Economic Data.
\newblock \url{https://fred.stlouisfed.org}.
\newblock Series SP500, DGS10, DEXUSEU, DHHNGSP, CPIAUCSL, INDPRO, UNRATE,
  VIXCLS, CBBTCUSD; accessed July 2026.

\bibitem[{Fu et~al.(2023)Fu, Dao, Saab, Thomas, Rudra, and
  R{\'e}}]{fu2023hungry}
Fu, D.~Y.; Dao, T.; Saab, K.~K.; Thomas, A.~W.; Rudra, A.; and R{\'e}, C. 2023.
\newblock Hungry Hungry Hippos: Towards Language Modeling with State Space
  Models.
\newblock In \emph{The Eleventh International Conference on Learning
  Representations}.

\bibitem[{Ganguli, Huh, and Sompolinsky(2008)}]{Ganguli2008}
Ganguli, S.; Huh, D.; and Sompolinsky, H. 2008.
\newblock Memory traces in dynamical systems.
\newblock \emph{Proceedings of the National Academy of Sciences}, 105(48):
  18970–18975.

\bibitem[{Gissin, Shalev-Shwartz, and Daniely(2020)}]{gissin2020the}
Gissin, D.; Shalev-Shwartz, S.; and Daniely, A. 2020.
\newblock The Implicit Bias of Depth: How Incremental Learning Drives
  Generalization.
\newblock In \emph{International Conference on Learning Representations}.

\bibitem[{Gu and Dao(2023)}]{gu2023mamba}
Gu, A.; and Dao, T. 2023.
\newblock Mamba: Linear-time sequence modeling with selective state spaces.
\newblock \emph{arXiv preprint arXiv:2312.00752}.

\bibitem[{Gu et~al.(2022)Gu, Goel, Gupta, and Ré}]{Gu2022}
Gu, A.; Goel, K.; Gupta, A.; and Ré, C. 2022.
\newblock On the Parameterization and Initialization of Diagonal State Space
  Models.
\newblock In \emph{Advances in Neural Information Processing Systems 35},
  NeurIPS 2022, 35971–35983. Neural Information Processing Systems
  Foundation, Inc. (NeurIPS).

\bibitem[{Gu, Goel, and Re(2022)}]{gu2022efficiently}
Gu, A.; Goel, K.; and Re, C. 2022.
\newblock Efficiently Modeling Long Sequences with Structured State Spaces.
\newblock In \emph{International Conference on Learning Representations}.

\bibitem[{Harrigan et~al.(2020)Harrigan, Zsoter, Alfieri
  et~al.}]{harrigan2020glofas}
Harrigan, S.; Zsoter, E.; Alfieri, L.; et~al. 2020.
\newblock {GloFAS-ERA5} Operational Global River Discharge Reanalysis
  1979--Present.
\newblock \emph{Earth System Science Data}, 12: 2043--2060.

\bibitem[{Hersbach et~al.(2020)Hersbach, Bell, Berrisford
  et~al.}]{hersbach2020era5}
Hersbach, H.; Bell, B.; Berrisford, P.; et~al. 2020.
\newblock The {ERA5} Global Reanalysis.
\newblock \emph{Quarterly Journal of the Royal Meteorological Society},
  146(730): 1999--2049.

\bibitem[{Huang et~al.(2017)Huang, Thorne, Banzon et~al.}]{huang2017ersst}
Huang, B.; Thorne, P.~W.; Banzon, V.~F.; et~al. 2017.
\newblock Extended Reconstructed Sea Surface Temperature, Version 5
  ({ERSSTv5}).
\newblock \emph{Journal of Climate}, 30: 8179--8205.

\bibitem[{Inness et~al.(2019)Inness, Ades, Agust{\'i}-Panareda
  et~al.}]{inness2019cams}
Inness, A.; Ades, M.; Agust{\'i}-Panareda, A.; et~al. 2019.
\newblock The {CAMS} Reanalysis of Atmospheric Composition.
\newblock \emph{Atmospheric Chemistry and Physics}, 19: 3515--3556.

\bibitem[{Jaeger(2002)}]{jaeger:techreport2002}
Jaeger, H. 2002.
\newblock Short term memory in echo state networks.
\newblock GMD-Report 152, GMD - German National Research Institute for Computer
  Science.

\bibitem[{Kailath, Sayed, and Hassibi(2000)}]{kailath2000linear}
Kailath, T.; Sayed, A.~H.; and Hassibi, B. 2000.
\newblock \emph{Linear Estimation}.
\newblock Upper Saddle River, NJ: Prentice Hall.

\bibitem[{Kalman(1960)}]{Kalman1960}
Kalman, R.~E. 1960.
\newblock A New Approach to Linear Filtering and Prediction Problems.
\newblock \emph{Journal of Basic Engineering}, 82(1): 35–45.

\bibitem[{{NOAA Global Monitoring Laboratory}(2026)}]{noaagml2026}
{NOAA Global Monitoring Laboratory}. 2026.
\newblock Mauna Loa {CO$_2$} Monthly Mean Data.
\newblock \url{https://gml.noaa.gov}.
\newblock Accessed July 2026.

\bibitem[{Orvieto et~al.(2023)Orvieto, Smith, Gu, Fernando, Gulcehre, Pascanu,
  and De}]{orvieto2023resurrecting}
Orvieto, A.; Smith, S.~L.; Gu, A.; Fernando, A.; Gulcehre, C.; Pascanu, R.; and
  De, S. 2023.
\newblock Resurrecting recurrent neural networks for long sequences.
\newblock In \emph{International conference on machine learning}, 26670--26698.
  PMLR.

\bibitem[{Pesme and Flammarion(2023)}]{NEURIPS2023_17a9ab41}
Pesme, S.; and Flammarion, N. 2023.
\newblock Saddle-to-Saddle Dynamics in Diagonal Linear Networks.
\newblock In Oh, A.; Naumann, T.; Globerson, A.; Saenko, K.; Hardt, M.; and
  Levine, S., eds., \emph{Advances in Neural Information Processing Systems},
  volume~36, 7475--7505. Curran Associates, Inc.

\bibitem[{Proca et~al.(2025)Proca, Domin{\'e}, Shanahan, and
  Mediano}]{proca2025learning}
Proca, A.~M.; Domin{\'e}, C. C.~J.; Shanahan, M.; and Mediano, P. A.~M. 2025.
\newblock Learning Dynamics in Linear Recurrent Neural Networks.
\newblock In \emph{Proceedings of the 42nd International Conference on Machine
  Learning (ICML)}.

\bibitem[{Roesler(2013)}]{roesler2013eeg}
Roesler, O. 2013.
\newblock {EEG} Eye State Data Set.
\newblock UCI Machine Learning Repository.

\bibitem[{Saxe, McClelland, and Ganguli(2014)}]{saxe2014exact}
Saxe, A.~M.; McClelland, J.~L.; and Ganguli, S. 2014.
\newblock Exact Solutions to the Nonlinear Dynamics of Learning in Deep Linear
  Neural Networks.
\newblock In \emph{International Conference on Learning Representations
  (ICLR)}.
\newblock ArXiv:1312.6120.

\bibitem[{{SILSO World Data Center}(2026)}]{silso2026}
{SILSO World Data Center}. 2026.
\newblock The International Sunspot Number.
\newblock Royal Observatory of Belgium, on-line Sunspot Number catalogue.
\newblock Accessed July 2026.

\bibitem[{Sm{\'e}kal et~al.(2024)Sm{\'e}kal, Smith, Kleinman, Biderman, and
  Linderman}]{smekal2024towards}
Sm{\'e}kal, J.; Smith, J. T.~H.; Kleinman, M.; Biderman, D.; and Linderman,
  S.~W. 2024.
\newblock Towards a Theory of Learning Dynamics in Deep State Space Models.
\newblock \emph{arXiv preprint arXiv:2407.07279}.
\newblock Presented at ICML 2024.

\bibitem[{Srinivasan, Laughlin, and Dubs(1982)}]{Srinivasan1982}
Srinivasan, M.~V.; Laughlin, S.~B.; and Dubs, A. 1982.
\newblock Predictive coding: a fresh view of inhibition in the retina.
\newblock \emph{Proceedings of the Royal Society of London. Series B.
  Biological Sciences}, 216(1205): 427–459.

\bibitem[{{The mathlib Community}(2020)}]{mathlib2020}
{The mathlib Community}. 2020.
\newblock The {L}ean Mathematical Library.
\newblock In \emph{Proceedings of the 9th {ACM} {SIGPLAN} International
  Conference on Certified Programs and Proofs ({CPP} 2020)}, 367--381. ACM.
\newblock Version v4.31.0, commit fabf563a7c95.

\bibitem[{{U.S. Geological Survey}(2026)}]{usgs2026}
{U.S. Geological Survey}. 2026.
\newblock National Water Information System.
\newblock \url{https://waterservices.usgs.gov}.
\newblock Accessed July 2026.

\bibitem[{White, Lee, and Sompolinsky(2004)}]{White2004}
White, O.~L.; Lee, D.~D.; and Sompolinsky, H. 2004.
\newblock Short-Term Memory in Orthogonal Neural Networks.
\newblock \emph{Physical Review Letters}, 92(14).

\bibitem[{Wiener(1949)}]{Wiener1949}
Wiener, N. 1949.
\newblock \emph{Extrapolation, Interpolation, and Smoothing of Stationary Time
  Series: With Engineering Applications}.
\newblock The MIT Press.
\newblock ISBN 9780262257190.

\end{thebibliography}
\end{document}